\documentclass[twocolumn]{article}
\usepackage{amssymb}
\usepackage[utf8]{inputenc}
\usepackage[T1]{fontenc}
\usepackage{amsmath}
\usepackage{graphicx}
\usepackage{makecell}
\usepackage{multirow}
\usepackage{float}
\usepackage{cite}
\usepackage{algorithm}
\usepackage{algorithmic}
\graphicspath{ }
\usepackage{comment}
\usepackage{placeins}
\usepackage{hyperref}
\usepackage{booktabs}
\usepackage{makecell}
\usepackage[table]{xcolor}

\hypersetup{
    colorlinks=true,
    linkcolor=blue,
    filecolor=magenta,
    urlcolor=cyan,
    }

\title{Large-Scale Multi-omic Biosequence Transformers for Modeling Protein-Nucleic Acid Interactions}
\author{Sully F. Chen\thanks{Authors contributed equally} \thanks{Duke University School of Medicine, Durham, NC 27710, USA}, Robert J. Steele\footnotemark[1] \thanks{NYU Langone Health, New York, NY 10016, USA}, Glen M. Hocky \thanks{Department of Chemistry and Simons Center for Computational Physical Chemistry, New York University, New York, NY 10012, USA}, Beakal Lemeneh\thanks{NYU Langone Health, New York, NY 10016, USA}, \\
Shivanand P. Lad\thanks{Duke University School of Medicine, Department of Neurological Surgery, Durham, NC 27710, USA}, Eric K. Oermann\thanks{NYU Langone Health, Department of Neurological Surgery, New York, NY 10016, USA}}
\date{}

\begin{document}

\maketitle

\section*{Abstract}
The transformer architecture has revolutionized bioinformatics and driven progress in the understanding and prediction of the properties of biomolecules. To date, most biosequence transformers have been trained on single-omic data—either proteins or nucleic acids—and have seen incredible success in downstream tasks in each domain, with particularly noteworthy breakthroughs in protein structural modeling. However, single-omic pretraining limits the ability of these models to capture cross-modal interactions. Here we present OmniBioTE, the largest open-source multi-omic model trained on over 250 billion tokens of mixed protein and nucleic acid data. We show that despite only being trained on unlabeled sequence data, OmniBioTE learns joint representations mapping genes to their corresponding protein sequences. We further demonstrate that OmniBioTE achieves state-of-the-art results predicting the change in Gibbs free energy ($\Delta G$) of the binding interaction between a given nucleic acid and protein. Remarkably, we show that multi-omic biosequence transformers \textit{emergently} learn useful structural information without any \textit{a priori} structural training, allowing us to predict which protein residues are most involved in the protein–nucleic acid binding interaction. Compared to single-omic controls trained with identical compute, OmniBioTE also demonstrates superior performance-per-FLOP across both multi-omic and single-omic benchmarks. Together, these results highlight the power of a unified modeling approach for biological sequences and establish OmniBioTE as a foundation model for multi-omic discovery.

\section*{Introduction}
\noindent
It has long been a fundamental goal of bioinformatics to derive functional and structural insights directly from primary biomolecular sequences. High-throughput sequencing technologies now enable routine acquisition of vast quantities of nucleic acid and protein data, yet translating these linear sequences into mechanistic understanding remains challenging. Recent breakthroughs in natural language processing (NLP), particularly the transformer architecture \cite{transformer}, have demonstrated exceptional capacity to model complex sequential dependencies in text. Despite these advances, cellular biology is inherently multi-omic, with proteins and nucleic acids engaging in dynamic and reciprocal interactions underpinning gene regulation, replication, and repair. Single-omic transformers, by design, lack the capacity to capture cross-modal dependencies in their fundamental representations to model tasks such as transcription factor binding, RNA-mediated translational control, and chromatin remodeling.

\begin{figure*}
\centering
\includegraphics[width=1.0\textwidth]{"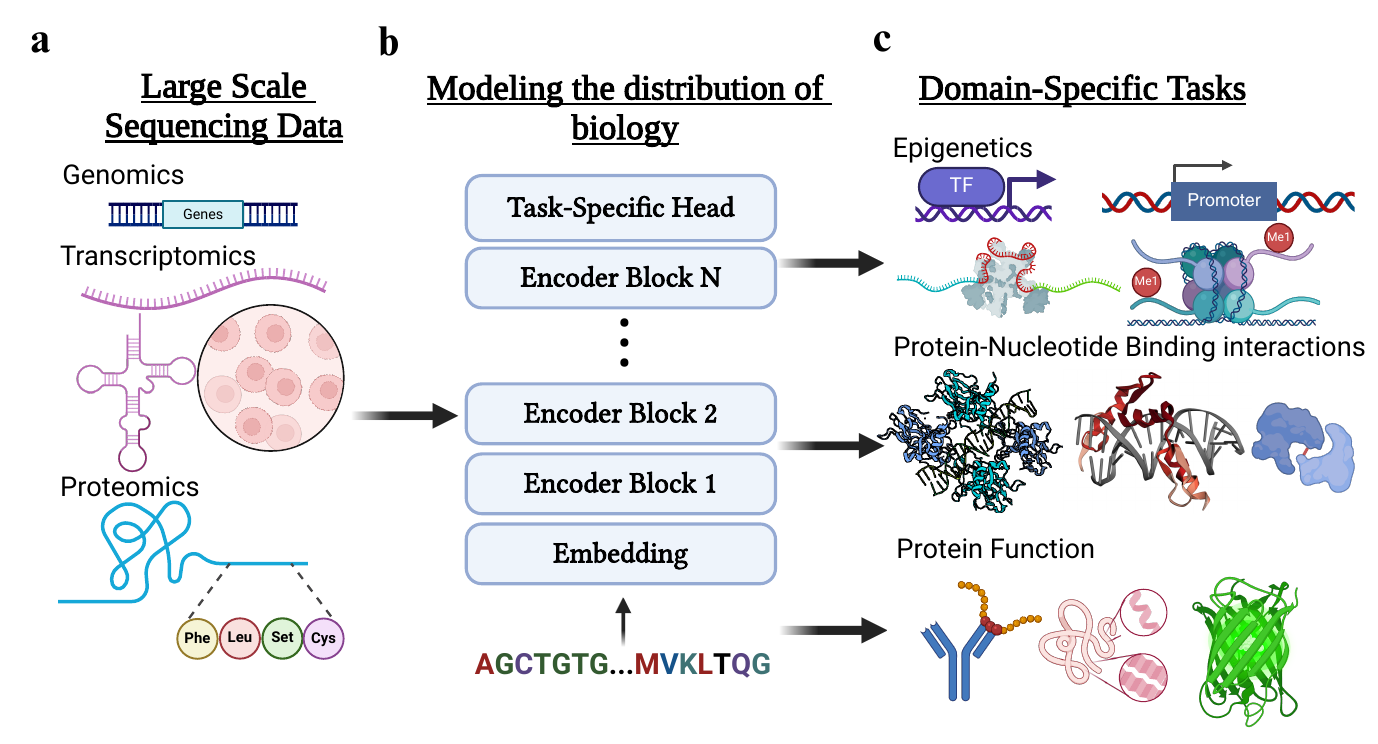"}
\caption{\textbf{Multi-omic pretraining and task-specific fine-tuning.} (\textbf{A}) First, we gather large-scale datasets consisting of proteomic data, nucleic acid modalities as DNA, many types of RNA, synthetic constructs, and more. (\textbf{B}) Next, we employ large-scale pretraining over these sequences via an encoder transformer and the masked language-modeling objective. (\textbf{C}) Finally, we fine-tune this foundation model with a task-specific head to tackle a wide variety of tasks. Created in BioRender. Chen, S. (2025) \url{https://BioRender.com/ydhbam8}}
\label{fig:overview}
\end{figure*}

Here, we introduce the OmniBioTE series of models and the first exploration of scaling laws in multi-omic transformers. Additionally, we contribute the largest open-source multi-omic transformer, pretrained on 250 billion tokens drawn from GenBank nucleic-acid entries and UniRef100 protein sequences (Fig~\ref{fig:overview}). We explore four model sizes (88M–2.3B parameters) and compare performance against matched single-omic controls (NucBioTE, ProtBioTE) trained with identical compute, but only nucleic acid data (NucBioTE) or on proteomic data (ProtBioTE). Notably, because total token budgets were fixed, each single-omic control is exposed to more unique single-omic data than the multi-omic model. We train four additional models that operate at the per-residue/nucleotide level (as opposed to tokenized chunks) to investigate the effects of tokenization on task-specific performance. For the multi-omic models, sequences of different modalities were never concatenated within the same context window during pre‑training, so no cross‑attention occurred across protein and nucleic‑acid tokens at pre‑training time. We evaluate on tasks spanning: (1) predicting binding free energies ($\Delta G$) for protein–nucleic acid complexes on ProNAB  \cite{10.1093/nar/gkab848}, (2) emergent contact prediction via attention-based probing, (3) nucleic acid specificity assessment on JASPAR \cite{JASPAR}, and (4) state-of-the-art performance on standard single-omic benchmarks (GUE \cite{zhou2024dnabert2efficientfoundationmodel}, TAPE \cite{tape}). Our results demonstrate that multi-omic pretraining yields embeddings that inherently align gene and protein modalities, outperform single-omic models in both multi-omic and single-omic tasks, and exhibit emergent structural knowledge without explicit supervision. OmniBioTE sets a new paradigm for foundation modeling in biology by unifying sequence modalities within a single transformer framework.

Our main contributions are as follows: we introduce OmniBioTE, a family of open-source multi-omic encoder transformers (88M–2.3B parameters; BPE and per-residue variants) jointly pretrained on 250 billion nucleic acid and protein tokens from GenBank and UniRef100, and release all models and code with a permissive open-source license. Second, we show that OmniBioTE learns modality-invariant gene–protein representations. Third, we develop multi-omic protein–nucleic acid interaction benchmarks—including a rigorously homology-filtered $\Delta G$ regression task on ProNAB, JASPAR-based mutation scans, and PDB-derived contact prediction—and demonstrate that OmniBioTE outperforms single-omic baselines, specialized models (DeePNAP), and an AlphaFold3-plus–molecular dynamics pipeline. Fourth, we demonstrate that the attention maps of our multi-omic models trained to predict binding energy emergently encode latent structural information. Finally, we perform a comprehensive scaling study on GUE, TAPE, and ProteinGLUE, showing that multi-omic pretraining improves performance per FLOP and establishes a new compute Pareto frontier on many single-omic tasks, despite reduced per-modality data.

\section*{Related work}
\noindent
The majority of research applying transformers to biosequences has focused on applying the architecture to single-omics, typically nucleic acid distributions (genomics, transcriptomics, epigenetics, etc.) or proteomics. These efforts have yielded astonishing successes in several tasks, with the most notable being the prediction of the 3D structure of proteins from their primary sequences \cite{alphafold1, alphafold2, alphafold3, rosettafold, rosettafoldna, openfold, omegafold, esmfold, drori2019accurate}. Other work has focused on developing models that produce useful representations of single-omics biosequences for various downstream tasks. There exist numerous protein foundation models \cite{elnaggar2023ankh, geffen2022distilprotbert, proberta, prottrans, rives2021biological, meier2021language, rao2021msa, prostt5, xtrimo100b, su2023saprot, notin2022tranception, alley2019unified, bepler2019learning}, and we find the most variety of model architectures in this class. Notably, there are many generative models \cite{alamdari2023protein, madani2023large, ferruz2022protgpt2}, encoder–decoder models \cite{prostt5, xtrimo100b}, and even a diffusion model \cite{alamdari2023protein}.

Several genomics foundation models have been trained as well, primarily on human genomics data \cite{genalm, hyenadna, dalla2023nucleotide, puffin}. Other genomic foundation models have been trained on human and murine data \cite{enformer}, multi-species genomes \cite{zhou2024dnabert2efficientfoundationmodel}, prokaryotic genomes \cite{genslm}, and even metagenomic scaffolds \cite{hwang2024}. Notably, very few models integrate broad, multi-species training data, with the exception of DNABERT-2 \cite{zhou2024dnabert2efficientfoundationmodel}, though this dataset notably lacks genomes from the domain Archaea and consists of only 32 billion nucleotides. To date, the largest DNA foundation model to be trained consists of 40 billion parameters \cite{evo2}, and was trained on multi-species genomes and found to be successful at multiple downstream tasks. Genomic models augmented with epigenetic data have also demonstrated great success in downstream tasks such as predicting epigenetic markers \cite{bert6ma, cpgtransformer, idnaabf, zhou2022deep}, detecting splice sites and promoter regions \cite{puffin}, modeling the histone code \cite{chromoformer}, and modeling the phosphorylation of protein kinases \cite{phosphormer}.

Other foundation models focus on transcriptomics, primarily focusing on single-cell RNA (scRNA) \cite{fu2023get, yang2022scbert, hao2024large, cui2024scgpt, theodoris2023transfer}. Other foundation models for mRNA \cite{codonbert} and general RNA \cite{bigrna} have also been trained. Transcriptomic foundation models have successfully predicted transcriptome-to-proteome translations, \cite{liu2023pre} gene rankings \cite{shen2023generative}, cell type annotation \cite{gong2024xtrimogene}, and drug response \cite{gong2024xtrimogene, theodoris2023transfer}.

Only three existing models incorporate both nucleic acid and protein information in a unified framework: AlphaFold3 \cite{alphafold3}, a closed-source proprietary model; RoseTTAFoldNA \cite{rosettafoldna}; and LucaOne \cite{lucaone}. The former two models are focused primarily on structure prediction rather than generally learning from multi-omic sequences, while the latter model’s nucleic acid sources are primarily sourced from RefSeq \cite{refseq}. RefSeq provides a sparse, curated subset: a single representative genome per organism, a reduced set of mature transcript and protein models, and virtually none of the underlying high-throughput data such as partial transcripts, genomic survey reads, metagenomic contigs, rare isoforms, immune V(D)J recombination products, or engineered sequences \cite{pruitt2020refseq}. As a result, large classes of biologically meaningful variation and sequence diversity present in GenBank are absent from RefSeq, potentially making it challenging for the model to learn robust representations of these classes (e.g., immunoglobulins or T-cell receptors). Furthermore, the largest open-source multi-omic model to date is LucaOne, with 1.8 billion parameters. In this work, we train a 2.3 billion parameter model, nearly 28\% larger. None of these models are open-source multi-omic sequence encoders trained at the scale and breadth of OmniBioTE, nor do they systematically study multi-omic scaling behavior across both single-omic and explicitly multi-omic benchmarks. A summary of the sizes of the models evaluated in this work can be found in Table \ref{tab:model_params}.

\begin{table}[!ht]
\small
\centering
\caption{Parameter counts for \textbf{OmniBioTE} and the external models evaluated in this work.}
\label{tab:model_params}
\begin{tabular}{lll}
\hline
\textbf{Family} & Variant & \textbf{Parameters} \\
\midrule
\multirow{4}{*}{\textbf{OmniBioTE} (this work)} 
  & Small  & 88M \\
  & Medium & 675M \\
  & Large  & 1.3B \\
  & XL     & 2.3B \\
\midrule
\textbf{LucaOne\cite{lucaone}} & -- & 1.8B \\
\midrule
\multirow{5}{*}{\textbf{ESM2\cite{esmfold}}} 
  & ESM2-XS   & 8M \\
  & ESM2-S   & 35M \\
  & ESM2-M  & 150M \\
  & ESM2-L  & 650M \\
  & ESM2-XL   & 3B \\
\midrule
\textbf{DNABERT-2\cite{zhou2024dnabert2efficientfoundationmodel}} & -- & 117M \\
\textbf{NT-2500M-multi\cite{dalla2023nucleotide}} & -- & 2.54B \\
\textbf{RandomMask\cite{randommask}} & -- & 110M \\
\textbf{DeePNAP\cite{pandey2024deepnap}} & -- & 173K \\
\textbf{AlphaFold3\cite{alphafold3}} & -- & closed-source \\
\bottomrule
\end{tabular}
\end{table}

\section*{Methods}
Broadly, we train dense, non-causal encoder transformer models of varying sizes using the masked-language-modeling (MLM) objective \cite{devlin2019bertpretrainingdeepbidirectional} on 250 billion tokens of nucleic acid and protein sequences of varying types. We additionally train control models consisting of only nucleic acid or protein sequences with equal compute budgets to evaluate the effect of training on additional sequence types. We demonstrate that our MOMs \textit{emergently} learn joint representations between nucleic acid and protein sequences by showing that there exist meaningful features roughly invariant to sequence modality, and that such features \textit{do not} exist in single-omic models.

We evaluate our suite of models by fine-tuning on several single-omics datasets that assess performance on various downstream tasks relevant to molecular biology, structural biology, and biochemistry. Additionally, we design two novel multi-omic tasks that require inference on both protein and nucleotide sequences simultaneously. Lastly, we show via simple convolutional probes that the models' attention maps encode structural information that is learned without any \textit{a priori} structural training.

\subsection*{Training Data}
We source our nucleic acid data from GenBank \cite{10.1093/nar/gky989}, a collection compiled by the National Center for Biotechnology Information. We preprocessed the entire GenBank archive by first removing all metadata from each sequence, with the exception of sequence type (DNA, mRNA, tRNA, etc.). This produced 242,855,368 sequences with a total of 312,190,748,151 nucleotides, primarily composed of general DNA, general RNA, mRNA, cRNA, and single-stranded RNA. A full breakdown of nucleic acid sequence data can be found in \nameref{dataset_statistics}. We source our protein data from UniRef100 \cite{10.1093/bioinformatics/btm098}, a dataset maintained by UniProt. Similarly to the nucleic acid data, we remove all metadata from each sequence, yielding 369,597,671 sequences with a total of 1,739,747,047 residues.

We take a subset of $10^{5}$ nucleotides and protein residues total to train a byte-pair encoding tokenizer \cite{sennrich2016neuralmachinetranslationrare} using the Sentencepiece library \cite{kudo2018sentencepiecesimplelanguageindependent}, with a vocabulary size of $2^{11}$ for protein sequences and nucleic acid sequences ($2^{12}$ unique tokens total since the vocabularies are disjoint). Our choice of tokenizer and vocabulary size was chosen based on previous work \cite{zhou2024dnabert2efficientfoundationmodel}. Additionally, we train a multi-omic per-residue/nucleotide model at each size to investigate the effects of tokenization on downstream performance, where each token is simply a single base pair or residue. In each case, we use a separate tokenizer for protein sequences and nucleic acid sequences. For example, the sequence ``ACGT'' is both a valid nucleic acid and peptide, and its tokenized representation will be different depending on the modality.

\subsection*{Architecture and Training}
OmniBioTE is based on the LLaMA-2 architecture \cite{touvron2023llama2openfoundation} with minimal modifications: we substitute learned positional embeddings \cite{transformer} with rotary positional embeddings (RoPE) \cite{su2023roformerenhancedtransformerrotary} and replace the causal self-attention mechanism \cite{transformer, radford2019language} with a full, non-causal attention operation \cite{devlin2019bertpretrainingdeepbidirectional}. We additionally scale the pre-SoftMax attention scores at $1/\text{width}$ rather than $1/\text{width}^2$ in accordance with maximal update parameterization ($\mu P$) \cite{yang2022tensorprogramsvtuning}. We use an aspect ratio (the ratio of model width to depth) of 128. We modify Karpathy's NanoGPT \cite{Karpathy2022} for a lightweight and simple model implementation. For a detailed description of the architecture, see \nameref{S1_Appendix}. We train four OmniBioTE variants, OmniBioTE-small (88 million non-embedding parameters), OmniBioTE-medium (675 million), OmniBioTE-large (1.3 billion) and OmniBioTE-XL (2.3 billion). Additionally, we train controls for each model on only nucleic acid data or only protein data (henceforth referred to as ``NucBioTE-[size]'' and ``ProtBioTE-[size]''). For experiments requiring fine-grained, single-nucleotide/residue inference, we also train an OmniBioTE model of each size that uses a single-character tokenizer rather than a byte-pair encoding (BPE). In total, we train 16 models: OmniBioTE-small/medium/large/XL, OmniBioTE (single-char)-small/medium/large/XL, ProtBioTE-small/medium/large/XL, and NucBioTE-small/medium/large/XL. Notably, the single-omic models and the multi-omic models have the same token budget, but different data mixtures. Thus, each single-omic model is trained on more unique data for its respective modality than the multi-omic models are.

We train each model for 250 billion tokens with a context length of 1024 tokens for the BPE-tokenized models and a context length of 2048 characters for the single-character models (to accommodate the decreased amount of data per token). We train at a batch size of 786432, 1032192, or 1048576 tokens (chosen based on available compute and memory and to maximize throughput) with the masked language modeling objective \cite{devlin2019bertpretrainingdeepbidirectional}. We use AdamW \cite{loshchilov2019decoupledweightdecayregularization} ($\beta_1 = 0.9$, $\beta_2=0.95$, $\epsilon=10^{-8}$, weight decay = $10^{-2}$), employing $\mu P$ for stable hyperparameter transfer. For the parameters with fixed learning rate under $\mu P$ (the embedding and unembedding parameters), we set the learning rate to $0.05$, and scale learning rates of the rest of the parameters via $32 / \text{width}$. These hyperparameters were determined empirically with sweeps at the $10^6$-parameter-scale. Finally, all learning rates are decayed with PyTorch's OneCycleLR \cite{smith2018superconvergencefasttrainingneural}, with a warmup period of 1 billion tokens, a starting and ending learning rate scale of $10^{-5}$.

\subsection*{Evaluations}
We design our own multi-omic benchmark to assess our model's ability to accurately characterize protein-nucleic acid interactions. We further design several novel benchmarks to assess the performance and interpretability of our models on protein-nucleic acid tasks. In addition to our main multi-omic tasks, we evaluate our approach on several popular benchmarks to evaluate single-omic performance on a variety of nucleic acid and protein-based tasks in an effort to assess the baseline single-omic capabilities of our model before multi-omic task-specific fine-tuning. All fine-tuning optimization is performed via AdamW \cite{loshchilov2019decoupledweightdecayregularization} with identical hyperparameters as described in the pretraining step unless otherwise specified.

\subsubsection*{Protein-Nucleic Acid Binding Evaluation}
To showcase the native multimodality of our generalist model, we designed a novel evaluation task using the ProNAB dataset \cite{10.1093/nar/gkab848}. ProNAB consists of 20,090 samples comprised of 14606 protein-DNA complexes, 5323 protein-RNA complexes, and 161 protein-DNA-RNA complexes. These samples are composed of 798 unique DNA-binding proteins and 340 unique RNA-binding proteins. We refer to the original work for a detailed description of the dataset composition \cite{10.1093/nar/gkab848}. The objective of our task is as follows: given the primary sequence of a nucleic acid-binding protein and a nucleic acid sequence, predict the $\Delta G$ of the binding interaction. This task is of particular interest in the prediction of unknown DNA/RNA-binding protein interactions with the human genome. 

We assemble our dataset by first filtering the ProNAB dataset, rejecting any nucleic acid or protein sequences with non-standard residues (we use only the standard 20 amino acids and the 5 standard nucleotide bases), leaving 850 unique proteins, and 15994 protein-nucleic acid complexes. We then split the data into 10 cross-validation sets. Ultimately, we end up with 752 unique proteins and 12282 total protein-nucleic acid interactions.

The ProNAB dataset often has multiple nucleic acid sequences per protein; thus the number of unique proteins is vastly outweighed by the number of unique nucleic acids. To avoid data leakage in the train and test sets, we group samples by protein sequence, then create folds by randomly grouping by protein sequence such that the folds do not have any proteins in common. Furthermore, we conduct sequence similarity analysis on the protein sequences in the train and test set via sequence alignment with the BLOSUM62 substitution matrix \cite{blosum62} to ensure minimal train/test leakage. We found that the average normalized alignment score between identical protein sequences in our dataset was $5.20 \pm 0.15$ (identical sequences may have different scores due to length normalization and BLOSUM62 scores), while over 99.4\% of pairwise comparisons in our train/test set had an alignment score below $0.0$, and 99.9\% had a score below $1.0$ suggesting that our results are not purely a result of sequence homology. As an extra precaution, we keep any proteins that have a sequence similarity score over $1.5$ \textit{with any other protein sequence in the dataset} strictly in the train set of all cross-validation sets to guarantee there is no significant sequence homology in any cross-validation fold. As a result, 13 unique proteins and 232 protein-nucleic acid interactions were always kept in the train set.

To compute a $\Delta G$ value, we first concatenate a primary protein sequence and nucleic acid sequence pair and run a forward pass through OmniBioTE. We then take the embedding produced by the first token and apply a linear projection which produces a single $\Delta G$ value. If a complex is composed of a protein and a double-stranded DNA or RNA molecule, we append the second nucleic acid sequence as well. We fine-tune our model to predict $\Delta G$ from the protein-nucleic acid pairs in the train set, with mean-squared error (MSE) as our loss target.  As a single-omic control, we compute the embeddings of the protein and nucleic acid sequences separately with the corresponding ProtBioTE and NucBioTE model. We then concatenate these embeddings and use a linear projection head to produce the $\Delta G$ value.

Our primary evaluation metrics are the Pearson correlation coefficient of $\Delta G$ prediction with the ground-truth measured value, as well as the mean absolute error of the predicted $\Delta G$ values. We begin with a pretrained OmniBioTE model, then further train our models for 64 epochs with a batch size of 256 on the $\Delta G$ prediction task. The projection head learning rate initialized to $10^{-2}$, the embedding vector learning rate initialized to $10^{-3}$, and the non-embedding parameters learning rate to $10^{-4} \cdot 1024/\text{width}$. All learning rates are decayed with PyTorch's OneCycleLR, an implementation of the learning rate schedule first described in \cite{smith2018superconvergencefasttrainingneural}.

As a baseline, we train a recent deep-learning-based architecture, DeePNAP \cite{pandey2024deepnap} on the identical cross-validation dataset as our model. We train the DeePNAP architecture for 64 epochs with a batch size of 256. For the training, we use AdamW ($\beta_1 = 0.9$, $\beta_2=0.999$, $\epsilon=10^{-8}$, weight decay = $10^{-2}$), starting at a learning rate of $10^{-3}$ and decaying linearly to $0.0$. Additionally, we fine-tune a recently released Genome-Protein model, LucaOne \cite{lucaone} in a similar manner. Specifically, we set the embedding learning rate to $10^{-4}$, the non-embedding parameter learning rates to $2.5\cdot10^{-5}$, and the projection head learning rate to $10^{-2}$. We train the LucaOne with identical AdamW hyperparameters, batch size, and epochs.

Lastly, we compare against a baseline that is more representative of current computational methods. First, we predict the structure of the protein-nucleic acid complex with AlphaFold3 \cite{alphafold3} and use molecular dynamics simulations to predict the $\Delta G$ of the binding interaction.

\subsubsection*{Nucleic Acid Binding Specificity}
To further validate the robustness of the OmniBioTE models fine-tuned to predict binding affinity, we evaluate whether the models can correctly predict the specificity of various DNA-binding proteins (DBPs) to their consensus sequences. First, we gather a set of 2,145 DBPs and their position-frequency matrices (PFMs) from JASPAR \cite{JASPAR}. Using the same sequence similarity rejection technique described in the ProNAB experiment, we filter all DBPs from the JASPAR dataset that have any significant overlap with the ProNAB dataset used in the cross-validation evaluation. We then use our fine-tuned OmniBioTE model to compute the $\Delta G$ for each DBP-nucleic-acid pair, where the consensus sequence is defined by the most frequent nucleotide in each position of the PFM. Next, we mutate each consensus sequence by randomly substituting each nucleotide with probability 5\%. This produces a mutated nucleic acid sequence that would have a reduced binding affinity to the DBP as empirically known by the PFM, but would still be ``in distribution'' of the plausible binding nucleic acids with high sequence similarity. We generate 8 unique mutated nucleic acid sequences per DBP. We predict the $\Delta G$ for these mutated interactions and compute the difference between the predicted $\Delta G$ of the consensus sequence. If the fine-tuned model has learned to model the specificity of the binding interaction correctly, we should expect the $\Delta G$ to increase after the consensus sequence is mutated.

\subsubsection*{Protein-Nucleotide Contact Prediction}
We gather all structures from the Research Collaboratory for Structural Bioinformatics Protein Data Bank \cite{10.1093/nar/28.1.235} that contain strictly one protein chain and either one or two nucleic acid chains. For each residue in the protein-nucleic acid complex, we compute the distance to the nearest nucleotide and label a residue as ``contacting a nucleotide'' if it is within a given distance threshold of a nucleotide. We test distance thresholds of 4~\AA{}, 6~\AA{}, and 8~\AA{}. Next, we group data by primary protein sequence and create 10 cross-validation splits by protein grouping to avoid data leakage. To fine-tune OmniBioTE, we concatenate the protein and nucleic acid sequences together and compute a forward pass through the model as usual. Instead of unembedding the hidden states of the final layers, we instead compute a linear projection to a single scalar, upon which a sigmoid function is applied to yield a contact prediction. Although the nucleic acid sequence is included in the forward pass, contact prediction is only computed for the protein residues. We train the model against a binary cross-entropy loss function for 32 epochs on each fold with a batch size of 256, with an identical training setup to the runs in the protein-nucleic acid binding evaluation. We additionally run the same training procedure on LucaOne with the embedding learning rate set to $10^{-4}$, the non-embedding parameter learning rates set to $2.5\cdot10^{-5}$, and the projection head learning rate set to $10^{-2}$, with identical AdamW hyperparameters.

\subsubsection*{Genome Understanding Evaluation} 
To evaluate OmniBioTE's generalizability to a variety of domain-specific nucleic acid tasks, we employ the Genome Understanding Evaluation (GUE) suite \cite{zhou2024dnabert2efficientfoundationmodel}. GUE consists of several genetic and epigenetic classification tasks over human, mouse, yeast, and coronaviridae genomes. Core promoter detection, transcription factor prediction, promoter detection, splice site detection, epigenetic mark prediction, and COVID variant classification were the target classes among these genomes. The promoter detection task is a binary classification task, where the goal is to determine whether a sequence of DNA is or is not a promoter. The promoter task is divided into several subcategories: proximal promoter detection, core promoter detection, and TATA/non-TATA motif promoter detection. The proximal promoter task contains the entire promoter sequence (including the core promoter) in the classification task, while the core promoter task only includes the sequence in close proximity to the transcription start site. The TATA class is composed of promoters that contain a TATA-motif, while the non-TATA does not have a TATA motif. Transcription factor detection is another binary classification task, where the goal is to determine whether a DNA sequence is the binding site of a transcription factor. This task is divided into human and murine datasets. Splice site detection is a classification task where the goal is to determine if a DNA sequence contains a splice donor or acceptor site. The epigenetic tasks' goals are to determine whether a nucleic acid sequence taken from a yeast genome is likely to contain a given epigenetic modification. Lastly, the COVID variant task is a multi-class classification task where the goal is to predict which variant type (Alpha, Beta, Delta, Eta, Gamma, Iota, Kappa, Lambda and Zeta) a 1000 base pair snippet was sequenced from. We refer to the original work for a full characterization of the evaluation set. All tasks use Matthews correlation coefficient as the primary metric, with the exception of the COVID variant classification task, which uses F1-score.

For each classification task, we fine-tune a base OmniBioTE or NucBioTE model. A class prediction is generated by taking the first token's final embedding and applying a linear projection down to the number of classes in place of the original final projection head, followed by a SoftMax operation. We set the embedding parameter learning rate to $10^{-3}$, the transformer weight matrices to $1024\cdot(\text{model width})^{-1}\cdot10^{-4}$, and lastly, set the learning rate of the projection head to $10^{-2}$ for all model sizes. Hyperparameters were determined with sweeps over the validation sets. All learning rates are decayed with PyTorch's OneCycleLR. The small and medium models are trained for 15000 steps with a batch size of 32 over the training data, while the large and XL models were trained for 30000 steps with a batch size of 32. We find that final validation performance is relatively robust to the number of epochs over each dataset, thus these training parameters were chosen to yield a reasonable training time. The model that performs best on the validation set is evaluated on the test set. We additionally fine-tune LucaOne as an additional multi-omic baseline. We train with the exact same optimizer hyper-parameters described for LucaOne in the protein-nucleic acid binding evaluation above. We train with batch size 32 for 30,000 iterations on each task.

\subsubsection*{Tasks Assessing Protein Embeddings} 
We employ the Tasks Assessing Protein Embeddings (TAPE) suite \cite{tape} to evaluate OmniBioTE's ability to generalize to unseen protein-based tasks. TAPE consists of five challenges: secondary structure prediction, residue contact prediction, remote homology detection, fluorescence prediction, and stability prediction. Secondary structure prediction is a per-residue classification challenge, where the goal is to determine what type of secondary structure each residue composes. The secondary structures are split into one of either 3 or 8 classes, depending on the task. Residue contact prediction involves generating an $N \times N$ mask, where $N$ is the length of the protein, with each element of the mask predicting the probability that a residue pair are within 8~\AA{} of each other. Remote homology detection involves mapping a primary protein sequence to one of 1195 homologies, with the aim to learn to classify primary sequences into meaningful structural families. Fluorescence prediction is a regression task, where the goal is to predict the log fluorescence intensity of a protein from a given primary structure. Finally, stability prediction is a regression task that aims to predict the maximum concentration at which a protein is still structurally stable. All classification tasks are measured in accuracy, while all regression tasks are measured via Spearman's correlation coefficient. We train each task (excluding the contact evaluation which is discussed below) for 64 epochs over the dataset with a batch size of 32, with identical initial learning rate parameters and schedule as the GUE tasks \cite{zhou2024dnabert2efficientfoundationmodel}, though we initialize the non-embedding model parameter learning rate to $1024\cdot(\text{model width})^{-1}\cdot10^{-4}$, the embedding learning rate to $10^{-4}$, and the projection head learning rate to $10^{-2}$ for all model sizes.

The residue contact evaluation task involves predicting an $L \times L$ matrix of values between $0$ and $1$, with each element $(i, j)$ representing the probability that residue $i$ in the primary sequence is within 8~\AA{} of residue $j$. To generate this prediction matrix, embeddings are generated from a transformer model \cite{transformer}, and a learned linear projection head transforms each embedding into $128$-dimensional vectors. As inspired by previous work \cite{ma2015protein}, a tensor of shape $256 \times L \times L$ is constructed, where item $[:, i, j]$ corresponds to the $i^{th}$ $128$-dimensional vector concatenated with the $j^{th}$ $128$-dimensional vector. This tensor is transformed via an 8-layer ResNet \cite{he2015deepresiduallearningimage} to yield a final $(1 \times L \times L)$ matrix, which after transformation by the sigmoid function, produces the desired probability matrix. Binary cross-entropy is used as the loss target, with masks applied computing the loss only on residue pairs that are separated by at least 12 total residues (excluding ``short'' contacts). Fine-tuning is performed for 128 epochs with a batch size of 128. The learning rate of non-embedding transformer parameters was set to $1024\cdot(\text{model width})^{-1}\cdot10^{-4}$, with the projection head and ResNet \cite{he2015deepresiduallearningimage} using a learning rate of $10^{-3}$. Learning rates were warmed up and decayed via the PyTorch OneCycleLR \cite{smith2018superconvergencefasttrainingneural} learning rate scheduler as mentioned previously.

We fine-tune a series of ESM2 models \cite{esmfold} to compare both absolute performance and scaling performance against a state-of-the-art single-omic protein model. Specifically, we fine-tune the 8 million, 35 million, 150 million, 650 million, and 3 billion parameter ESM2 models in an identical fashion as the OmniBioTE models above. For brevity, we hereafter refer to the ESM models as ESM2-XS (8 million), ESM2-S (35 million), ESM2-M (150 million), ESM2-L (650 million), and ESM2-XL (3 billion). We use the same embedding and head learning rate as the OmniBioTE finetuning runs, and set the non-embedding parameter learning rate to $640\cdot(\text{model width})^{-1}\cdot10^{-4}$. Additionally, we evaluate LucaOne via the same hyperparameters described in the protein-nucleic acid binding evaluation, with the same number of iterations and batch size for each task. We use AdamW ($\beta_1=0.9$, $\beta_2=0.999$, $\epsilon=10^{-8}$, weight decay = $0.01$) as the optimizer for all models.

\subsubsection*{Protein General Language of Life Evaluation} 
To explore per-residue tasks (i.e., tasks that require a prediction for every residue in the protein), we employ the Protein General Language of Life Evaluation (ProteinGLUE) \cite{Capel2021.12.13.472460}. We refer to the original work for a full description of ProteinGLUE, but briefly, ProteinGLUE consists of several tasks:

Secondary structure prediction: the task is identical to the TAPE secondary structure task discussed above \cite{tape}. Accuracy is the primary metric.

Solvent accessibility: the task is to either classify whether a residue has less than 7\% solvent accessibility, as well as a regression task to predict the actual solvent accessibility value. For the binary classification task, accuracy is the primary metric, and Pearson correlation coefficient is used as the primary metric for the regression task.

Protein-protein interaction: the task is to predict which residues interact in either homodimer or heterodimers. Area under the receiver operating characteristic curve (AUCROC) is used as the primary metric.

Epitope region detection: the task is to predict which regions of a protein are antigenic epitopes. The performance of this task is measured in AUCROC.

Hydrophobic patch prediction: the goal of this task is to predict the largest rank of a hydrophobic patch that a residue belongs to. This task is measured via Pearson correlation coefficient.

Each task was trained with a batch size of 32 for 16 epochs on all tasks except for the protein-protein interaction, for which 64 epochs were used owing to a smaller dataset size. Identical initial learning rates and schedules used in the TAPE evaluation mentioned above were used. We compare against ESM models in a similar manner as the TAPE evaluations, namely with an embedding learning rate of $10^{-4}$, a projection head learning rate of $10^{-2}$, and a non-embedding parameter learning rate of $640\cdot(\text{model width})^{-1}\cdot10^{-4}$. We use the same optimizers and hyperparameters as described in the TAPE evaluations. We evaluate LucaOne on this task with identical hyperparameters as the TAPE evaluation.

\subsection*{Per-Residue Evaluations}
Because the protein and nucleic acid datasets were tokenized with byte-pair encoding, most tokens contain several nucleotides or residues. Evaluations that require a per-residue prediction, such as secondary structure, are not directly compatible with this tokenization scheme. To solve this issue, we apply two simple strategies at train and test time. At train time, we compute the target of a single token as the mode of all the residues it contains in the case of a classification task or the mean of the values of the residues it contains in the case of a regression task. This allows the input sequence length and the target sequence length to be the same size. At test time, we simply duplicate the value at the predicted token by the number of residues that token contains, allowing us to construct a prediction with the same length as the target ground truth. This method places an upper bound on the maximum achievable performance our model can achieve on any per-residue task, but in practice, this upper bound is higher than state-of-the-art results previously reported. This is likely due to the fact that nearby residues often share similar values in per-residue prediction tasks (e.g., if a residue is in a beta chain, its adjacent residues are likely to be in a beta chain as well). We note that our evaluation results are still directly comparable to previous per-residue methods, as we duplicate our predictions to match the ground truth dimensionality rather than decreasing the ground truth dimensionality to match the sequence length (as is done at train time).

For the contact evaluations, the non-uniform number of residues encoded by each token presented a significant challenge. We remedy this issue by transforming prediction targets from residue to token space for training and transforming predictions from token to residue space for evaluation. Transformation of prediction maps from residue space to token space was accomplished by assigning the $(i, j)$-token pair as a true contact if \textit{any} of the residues contained within token $i$ contact \textit{any} of the residues within token $j$. Similarly, the $(i, j)$-token pair of the contact mask, used to ignore short-range contacts in the loss function, was assigned a positive value if any of the residues contained within token $i$ are at least 12 residues apart from any of the residues contained in token $j$. Transforming from token space to residue space for evaluation is done in a simpler manner: residue $(n, m)$ is assigned the value of the token pair $(i, j)$, where $i$ is the token containing residue $n$ and $j$ is the token containing residue $m$. For the per-residue/nucleotide models, the models were evaluated normally.

\subsection*{Interpretability}
\subsubsection*{Protein-Nucleic Acid Interactions}
To show that OmniBioTE learns semantically meaningful features, we demonstrate that when trained to predict the binding affinity between a nucleic acid and a protein sequence, OmniBioTE implicitly learns structural information despite exclusively being trained on primary sequence data. We fine-tune one OmniBioTE model of each size, in an identical fashion as described for the protein-nucleic acid binding evaluation, though we use all available data rather than cross-validation splits, as the goal is to fine-tune OmniBioTE models to be highly capable of predicting binding interactions, then investigate their mechanics.

Next, we gather all structures from the Research Collaboratory for Structural Bioinformatics Protein Data Bank \cite{10.1093/nar/28.1.235} that contain strictly one protein chain and either one or two nucleic acid chains. For each residue in the protein-nucleic acid complex, we classify the residue as making contact with a nucleotide if it is within 8~\AA{} of any nucleotide (in the same manner as described in the Protein-nucleic acid Contact Prediction task). We then compute a forward pass through either the OmniBioTE model fine-tuned to predict $\Delta G$ or through the base OmniBioTE model (control) and collect the attention maps produced by each head in each layer (this results in $N^2$ attention maps, where $N$ is the number of layers). Next, we concatenate these attention maps along the channel dimension to produce an $N^2\times L\times L$ tensor, where $L$ is the length of the input sequence. We then train a small convolutional network consisting of four layers. The first layer takes the $N^2$ channels and applies a $3\times 3$ convolution to produce 64 channels, the next two layers apply a $3\times 3$ convolution producing 64 channels, and the final layer again applies a $3 \times 3$ convolution but produces only one channel. The output of the convolutional net is an $L\times L$ tensor, and we average across the last dimension to produce $L$ logits that, after a sigmoid operation, yield the predicted probability that a given residue makes contact with a nucleotide (this task is identical to the Protein-Nucleic acid Contact Prediction task described above). We train this convolutional network via AdamW with a learning rate of $10^{-3}$, $\beta_1=0.9$, $\beta_2=0.999$, weight decay of $10^{-2}$, and $\epsilon=10^{-8}$ for 1000 steps with a batch size of 256, linearly decaying the learning rate to zero over the course of training. Critically, \textit{the weights of the underlying OmniBioTE model remain frozen throughout training}, meaning that the convolutional network must extract this structural information strictly from the attention maps produced by the underlying model. We compare the F1-score on each of the 10 folds for the attention maps produced by the base OmniBioTE model and those produced by the OmniBioTE model fine-tuned to predict binding affinity. If the fine-tuned model has learned meaningful structural information from the fine-tuning process, we would expect the F1-score for convolutional networks trained on these attention maps to be higher than those of the base model.

\subsubsection*{Shared Representations Between Modalities}
We aim to test whether OmniBioTE effectively learns a joint representation space between nucleic acid and protein sequences rather than simply learning to represent both modalities separately. In this case, we want to test whether OmniBioTE has learned representations of gene sequences (DNA, both coding and non-coding regions) and their corresponding protein sequences that reflect shared functional or structural properties, independent of the sequence modality.

We first formalize the notion of invariance under transcription and translation. Let \(x \in X\) be a gene (DNA) sequence, and let \(y \in Y\) be the corresponding protein sequence produced by a mapping \(G : X \to Y\), such as the standard transcription and translation process. Suppose that our pretrained multimodal model outputs embeddings \(\mathbf{z}_x\) for \(x\) and \(\mathbf{z}_y\) for \(y\), where \(\mathbf{z}_x, \mathbf{z}_y \in \mathbb{R}^d\). We define a feature extractor \(\phi: \mathbb{R}^d \to \mathbb{R}^k\) that maps an embedding to a scalar feature value. A feature is called \emph{invariant} under the mapping \(G\) if
\[
\phi(\mathbf{z}_x) = \phi(\mathbf{z}_y)
\]
for all \(x \in X\) and \(y = G(x)\). In practical terms, such an invariant feature may correspond to the biological function or identity of a gene–protein pair, that is, a characteristic that remains constant regardless of the modality.

To test whether the model has indeed learned such invariant features, we conduct a contrastive learning experiment employing a strict linear transformation. In this experiment, we first obtain pairs of gene sequences (including both intronic and exonic regions) and their corresponding translated protein sequences. Using our pretrained multimodal model, we compute the embeddings \(\mathbf{z}_x\) and \(\mathbf{z}_y\) for each gene and protein sequence, respectively. We then introduce a learnable linear transform \(W \in \mathbb{R}^{k \times d}\) with low rank \(k \ll d\) to project the embeddings into a shared subspace, yielding \(W \mathbf{z}_x\) and \(W \mathbf{z}_y\). The function \(W\) is optimized via a contrastive objective that simultaneously maximizes the cosine similarity between corresponding pairs \(W \mathbf{z}_x\) and \(W \mathbf{z}_y\) while minimizing the similarity between non-corresponding pairs.

Specifically, we employ a contrastive loss function similar to the CLIP framework \cite{clip} to  learn our feature extractor: let \(X \in \mathbb{R}^{N \times d}\) and \(Y \in \mathbb{R}^{N \times d}\) denote two batches of embeddings (with \(N\) samples and embedding dimension \(d\)), where each row $x_i$  of $X$ is a gene's feature vector and each row $y_i$ of $Y$ is the corresponding protein sequence. Any given pair $x_i$ and $y_j$ are unrelated if $i \neq j$. To compute the contrastive loss, each embedding in \(X\) and \(Y\) is normalized to unit length. The normalized embeddings are then used to compute a similarity matrix \(S \in \mathbb{R}^{N \times N}\) whose entries are given by
\[
S_{ij} = \frac{\langle \hat{x}_i, \hat{y}_j \rangle}{\tau},
\]
where \(\tau\) is a temperature parameter that controls the scaling of the cosine similarities.

In this setup, the diagonal elements \(S_{ii}\) represent the cosine similarity between corresponding pairs, while the off-diagonal elements \(S_{ij}\) for \(i \neq j\) represent the similarities between non-corresponding pairs. Our final loss is composed of two terms: the first term considers each row of \(S\) as logits for a classification task in which the correct label for \(x_i\) is \(i\). The second term is computed by treating each column as logits for the corresponding \(y_i\). The two terms are simply averaged to compute the final scalar loss. This approach is identical to the original CLIP loss proposed by Radford et al.\cite{clip}. For our experiments, we use $\tau=0.07$, and $k=16$.

We minimize this loss via the AdamW optimizer, with learning rate 0.01, linearly decayed to $0.0$ over $10000$ steps, $\beta=(0.9, 0.95)$, and $\epsilon=10^{-8}$. \textit{We optimize strictly over the projection matrix and leave the model parameters frozen}, as the goal is to test whether joint features are already learned, not whether they \textit{can} be learned.

After learning $\phi$, we apply this transformation to a held-out set of gene-protein pairs and compute the dot product between their feature representations. If $\phi$ is a generalizable feature extractor, we should see high dot product scores between corresponding held-out pairs and low dot product scores between non-corresponding held-out pairs.

Critically, we assess the generalization capability of the invariant features under very strict conditions; we train on only 5\% of the available paired data and test on the remaining 95\%. Strong performance in this setting indicates that the model’s embeddings encode a shared subspace that captures the desired invariances.

For further validation, we perform a control experiment using two separately trained single-omic models—one trained solely on genes and the other solely on proteins. In this case, the embedding spaces of these models are learned independently, and there is no inherent guarantee of alignment between them. We attempt to learn two distinct feature extractors, \(\phi_x\) and \(\phi_y\), for the gene and protein modalities, respectively, with the goal of minimizing the same contrastive loss.

\section*{Results}
\subsection*{Emergent joint representations}
\begin{figure*}
\centering
\includegraphics[width=1.0\textwidth]{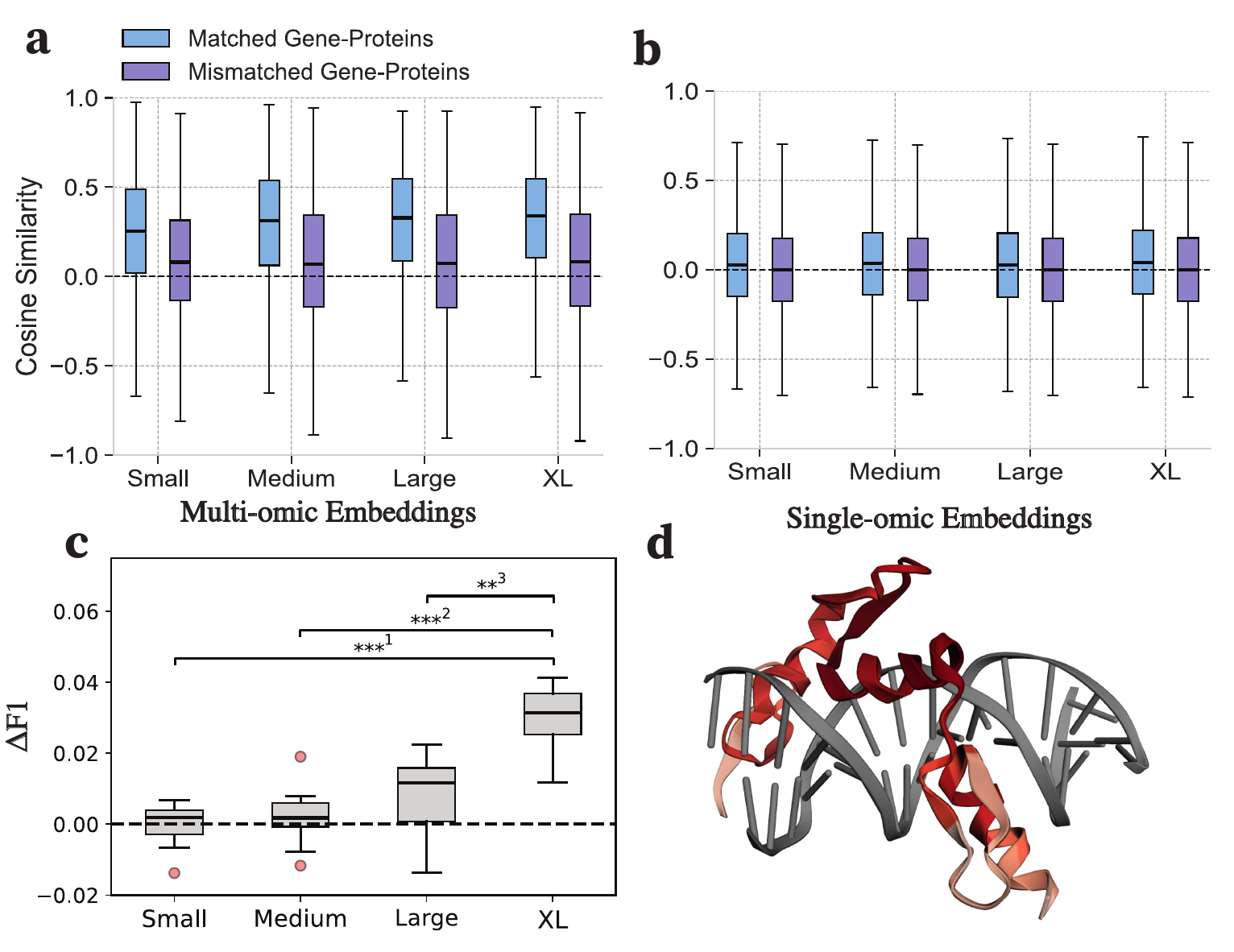}
\caption{\textbf{Emergent alignment of gene and protein embeddings and latent structural information.} (\textbf{a}) The distribution of cosine similarity between feature vectors produced by OmniBioTE via a low-rank feature extractor on the 95\% held-out data. (\textbf{b}) The analogous plot produced by NucBioTE and ProtBioTE with two separate feature extractors with identical methodology. (\textbf{c}) The increase in F1-score on the contact-prediction task using frozen attention maps from OmniBioTE models fine-tuned to predict binding affinity compared to frozen attention maps from the base models. (\textbf{d}) An example of predicted contact probability for zinc-finger and BTB domain-containing protein 7A (ZBTB7A) bound to a DNA duplex computed from the attention maps produced by the fine-tuned OmniBioTE models. Darker red colors indicate a stronger predicted probability of contact. All box-and-whisker plots are constructed via the median value as the central line, the interquartile range (IQR) as the box, and the whiskers denoting the minimum and maximum value of the distribution. Outliers are defined as points that lie outside of $\pm 1.5 \times \mathrm{IQR}$ and were excluded from (a) for clarity. $***^1$: $p=2.5\times10^{-6}$, $***^2$: $p=8.8\times10^{-6}$, $**^3$: $p=6.3\times10^{-4}$. P-values were computed via one-sided Welch's t-test with Holm-Bonferroni correction for multiple comparisons. Significance was determined at $\alpha=0.01$. Significance testing in (\textbf{a}) and (\textbf{b}) is omitted due to extremely large sample size leading to trivially high significance.}
\label{fig:interp}
\end{figure*}

We first tested whether OmniBioTE embeddings encode modality-invariant features linking genes and proteins. First, we generate embeddings for the primary sequences of a set of proteins, as well as the genes that encode them (both non-coding and coding regions). Next, a low-rank linear projector is trained on these frozen embeddings via a contrastive loss objective (with matching protein-gene pairs serving as positives) with only 5\% of ground-truth data. This simple linear probe is best thought of as a transform that narrows into a small subspace of the overall embedding space, rather than a feature extractor. Remarkably, we find that the contrastive performance of small linear probes trained on only 5\%  of the gene-protein pairs generalize well to the remaining 95\% of held-out data (Fig~\ref{fig:interp}a,b). In comparison, two separate low-rank linear probes trained with identical objectives and data splits on the single-omic models fail to generalize. Despite OmniBioTE never being explicitly (or even implicitly) taught a correspondence between genes and their corresponding translated protein sequences, the model naturally learns these associations from the underlying distributions. Furthermore, the failure of single-omic models to generalize despite loosening constraints to two separate linear probes demonstrates that the generalization is due to joint embeddings, rather than matching corresponding extracted features.

\subsection*{Performance on multi-omic tasks}
\begin{figure*}
\centering
\includegraphics[width=1.0\textwidth]{"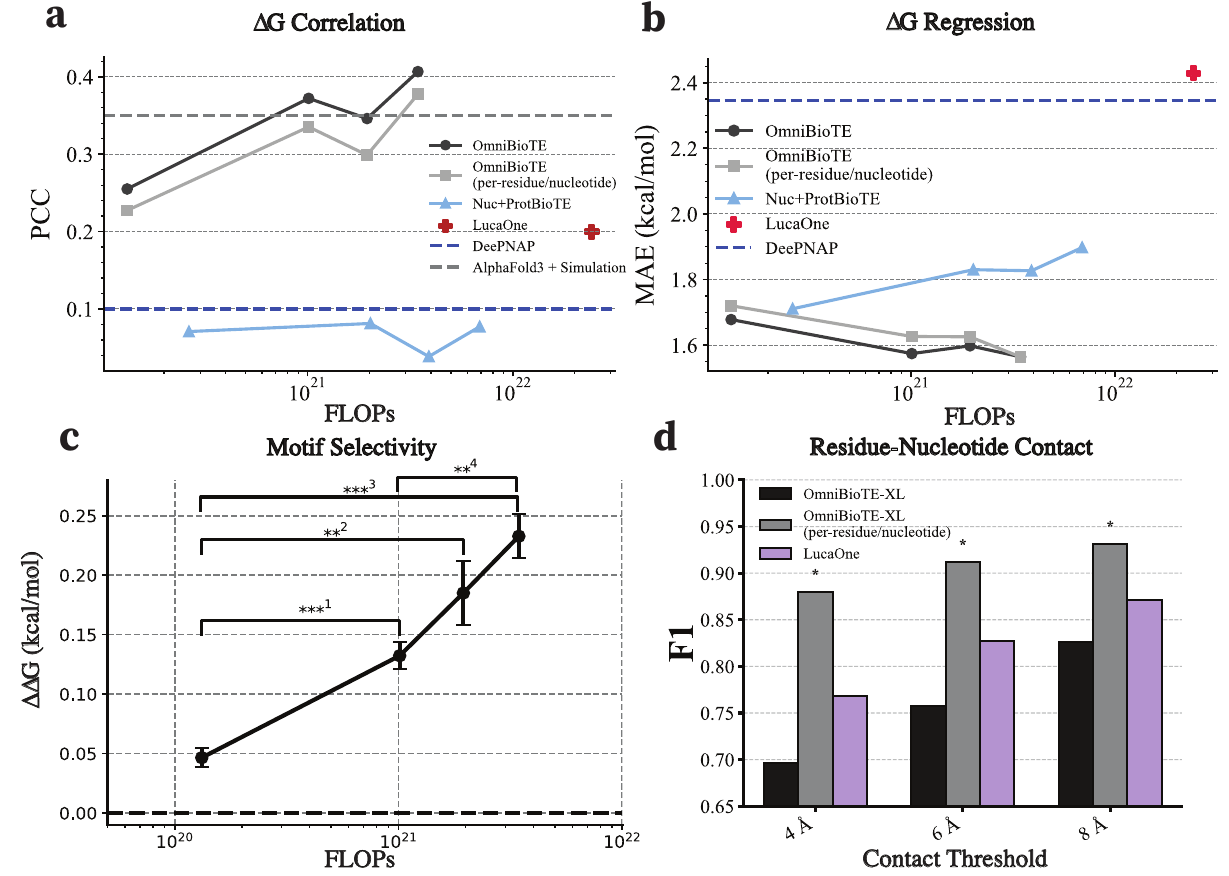"}
\caption{\textbf{Multi-omic pretraining facilitates state-of-the-art results on protein-nucleic acid complex $\Delta G$ regression.} (\textbf{A}) Performance on 10-fold cross-validation over the ProNAB dataset as measured by the Pearson correlation coefficient (PCC) as a function of pretraining compute. (\textbf{B}) Mean absolute error in $\Delta G$ prediction over the 10-fold cross-validation set. (\textbf{C}) The predicted $\Delta\Delta G$ of mutated consensus sequences as a function of pretraining compute. Error bars represent the standard error of the mean of all 10 folds. LucaOne and DeePNAP baselines are omitted for clarity, as both achieve performance similar to random chance ($\Delta\Delta G=0$). (\textbf{D}) Performance on the supervised contact evaluation task trained at various contact thresholds. The positive-to-negative ratio of the dataset is 0.29, 0.16, 0.09, and the maximum F1-score achievable with random guessing is 0.37, 0.247, and 0.157, for 8~\AA{}, 6~\AA{}, and 4~\AA{}, respectively. (*) represents the top-performing model in each evaluation. $***^1$: $p=6.7\times10^{-5}$, $**^2$: $p=1.5\times10^{-3}$, $***^3$: $p=4.3\times10^{-6}$, $**^4$: $p=1.3\times10^{-3}$. P-values were computed via one-sided Welch's t-test with Holm-Bonferroni correction for multiple comparisons. Significance was determined at $\alpha=0.01$.}
\label{fig:multiomics}
\end{figure*}

We demonstrated OmniBioTE's potential as a foundation model for natively multi-omic tasks by fine-tuning each OmniBioTE model to predict the $\Delta G$ of protein-nucleic acid binding interactions. We measured the Pearson correlation coefficient between the laboratory-measured $\Delta G$ value and the value predicted by OmniBioTE, as well as the mean absolute error between these values. We found that our largest model achieved a Pearson correlation coefficient of $0.41$ and MAE = 1.56 kcal/mol, exceeding single-omic controls ($\Delta$PCC=+0.33) (Fig~\ref{fig:multiomics}a,b). In addition to our single-omic controls, we compared against a recently developed binding affinity regression model, DeePNAP \cite{pandey2024deepnap}, as well as a computationally intense molecular dynamics-based simulation on structures predicted by AlphaFold3 \cite{alphafold3}. We find that after rigorously partitioning the train and test sets by sequence homology (via alignment scores generated with BLOSUM62 substitution matrices), our largest model considerably outperforms both DeePNAP and the AlphaFold3 + molecular dynamics predictions. AlphaFold3 based simulations were notably more computationally intensive (\nameref{S2_Appendix} and \nameref{S1_Fig}). The full results of the evaluation can be found in \nameref{pronab_results}. As a performance ceiling, we note that empirical work has found that the maximum possible Pearson's correlation coefficient is around 0.81, and the minimum possible mean absolute error is around 0.6 kcal/mol\cite{wetlabdeltaG}.

Next, we aimed to evaluate whether our fine-tuned models could model the specificity of DNA-binding proteins. We introduced small mutations into the consensus sequences of DNA-binding proteins from the JASPAR dataset \cite{JASPAR}, yielding highly similar sequences that should have strictly lower binding affinity. These mutation scans confirmed that $\Delta\Delta G$ predictions increase upon subtle consensus sequence disruption on average, scaling with model size (Fig~\ref{fig:multiomics}c). Additionally, we find that the increase in the magnitude of the predicted $\Delta\Delta G$ increases significantly with scale (Spearman's $\rho=0.86$, $p=1.16\times 10^{-12}$). This result demonstrates that our fine-tuned model is sensitive to fine-grained changes in sequences, rather than modeling a rough distribution of binding affinities belonging to families of proteins or consensus sequences. Furthermore, the generalization of our method to sequences from JASPAR, a dataset that is out-of-distribution with respect to the training set, demonstrates the validity and robustness of our methodology.

We found that the multi-omic approach is considerably more performant and compute-efficient than using two identically trained single-omic models (Fig~\ref{fig:multiomics}a,b). We found a clear trend of increasing performance with model scale, as opposed to over-fitting with greater parameter count, indicating the robustness of the approach and potential for further performance gains with greater scale in both compute and data.

As another multi-omic structural task, we fine-tuned OmniBioTE models to take as input the primary sequence of both the protein and nucleic acid in a given protein-nucleic acid complex (sourced from the RCSB Protein Data Bank \cite{10.1093/nar/28.1.235}) and predict which residues in the complex make contact with the nucleic acid it binds to, with contact defined as a residue and nucleotide residing within a given distance threshold. We find that on the protein-nucleic acid contact prediction task (measured in F1-score), our per-residue/nucleotide OmniBioTE-XL model outperforms a genomic/proteomic baseline, LucaOne, which had considerably more pretraining compute invested (Fig~\ref{fig:multiomics}d) and that results improve with model scale. We hypothesize that this advantage stems from training OmniBioTE on a wide variety of nucleic acid data, in addition to genomics. We find that the byte-pair encoded OmniBioTE model underperforms compared to the LucaOne baseline and the per-residue/nucleotide OmniBioTE models, which we attribute to lower-resolution predictions (each token predicts the contact for multiple residues at a time). Additionally, we find similar improvements with scale on the contact prediction task (\nameref{pdb_contact_f1}).

\subsection*{Attention-based Structural Interpretability}
We assessed whether attention maps extracted from OmniBioTE models fine-tuned to predict binding affinity implicitly encoded structural information, despite having no \emph{explicit} structural training data. A simple convolutional probe was trained on frozen attention maps from OmniBioTE models fine-tuned to predict binding affinity and compared to an identical convolutional probe trained on frozen attention maps produced by their corresponding base models. Critically, all model parameters were frozen while training the probes, ensuring that no structural information leaked into either model's attention maps. If the simple convolutional probes trained on frozen attention maps from the fine-tuned models consistently yield better prediction performance than identical probes fine-tuned on attention maps from base models, then it can be concluded that the fine-tuned model implicitly encodes richer structural information. We found that the probe trained on attention maps from the fine-tuned OmniBioTE models yielded consistently higher F1 scores on the contact prediction task at larger model scales (Fig~\ref{fig:interp}c), indicating that more latent structural information is present in the attention maps produced by models trained to predict binding affinity. This is particularly striking as this structural information is not explicitly present in the binding affinity task and must instead be inferred. Additionally, the difference in F1 score increases with model size (Spearman's $\rho=0.70$, $p=4.2\times10^{-7}$), suggesting that larger pretrained models may be better at inferring structural information. An example of contact predictions projected onto a zinc-finger protein is shown in Fig~\ref{fig:interp}d.

\subsection*{Performance on single-omic benchmarks}
\begin{figure*}
\centering
\includegraphics[width=1.0\textwidth]{"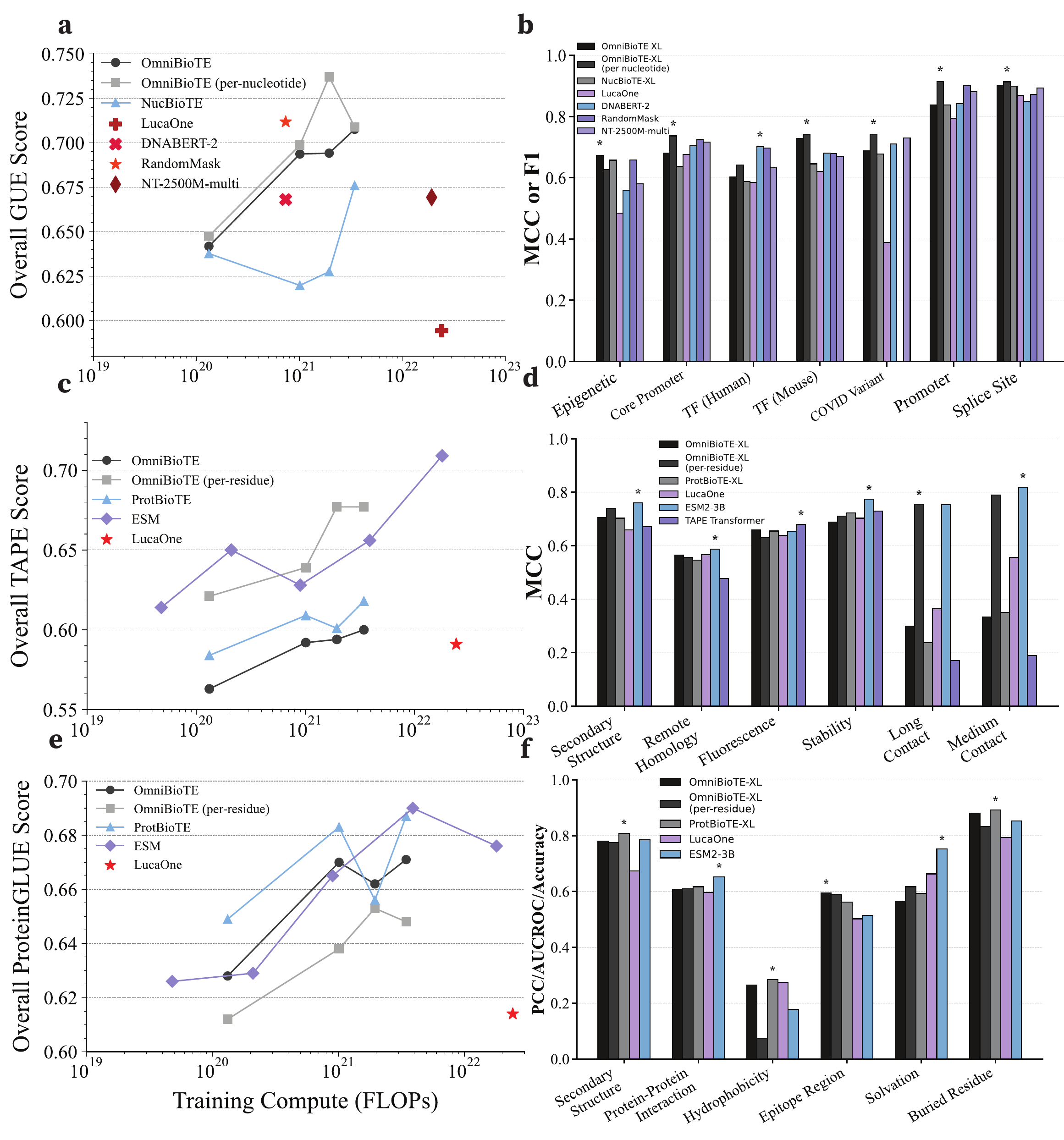"}
\caption{\textbf{Performance and scaling across single-omic benchmarks.} Aggregate benchmark performance for each model plotted as a function of pretraining FLOPs for the (\textbf{a, b}) GUE, (\textbf{c, d}) TAPE, and (\textbf{e, f}) ProteinGLUE benchmarks demonstrating superior performance per pretraining FLOPs of multi-omic pretraining compared to single-omic pretraining. GUE epigenetic mark prediction benchmarks were averaged to form a single category. Each point in the OmniBioTE series represents small, medium, large, and XL from least to most parameters in panels (a,c,e). (*) represents the top-performing model in each evaluation.}
\label{fig:single-omic}
\end{figure*}

We hypothesized that our multi-omic model may be more performant on single-omic benchmarks, especially from the perspective of performance-per-FLOP or performance per dataset size, two metrics that are broadly recognized as critical metrics for scaling large transformer models. For each benchmark across all tasks, multi-omic pretraining demonstrates superior or comparable performance to single-omic pretraining in terms of performance-per-FLOP even with vastly different compute budgets for the GUE, TAPE, and ProteinGLUE benchmarks (Fig~\ref{fig:single-omic}a,c,e). This improvement in performance-per-FLOP is even more striking when considering that significantly less data per-modality was seen by the model in the multi-omic training runs, since the total token budget was fixed in all training runs regardless of modality. In the GUE benchmarks (Fig~\ref{fig:single-omic}b), OmniBioTE models set a new state-of-the-art in all categories, with the exception of human transcription factor classification. All sizes of the OmniBioTE models lie well above the previous compute-to-performance Pareto frontier, with the exception of the RandomMask model, indicating strong scaling across over an order of magnitude of compute. The full results of the GUE evaluations can be found in \nameref{epigenetics1}, \nameref{epigenetics2}, \nameref{tf_covid}, \nameref{mouse_tf}, \nameref{promoter_detection}, and \nameref{core_promoter}. In the TAPE evaluations (Fig~\ref{fig:single-omic}d), OmniBioTE does not achieve any state-of-the-art results in terms of absolute performance, but the per-residue OmniBioTE models begin to trend above the previous compute Pareto frontier set by ESM, with only the smallest OmniBioTE model lying below the Pareto frontier. The results of the TAPE evaluation can be found in \nameref{tape1}, \nameref{tape2}, and \nameref{tape3}. Results are mixed between all models on ProteinGLUE (Fig~\ref{fig:single-omic}f), with the Pareto frontier difficult to ascertain; more scaling experiments are likely needed to elucidate the true frontier. The full results of the ProteinGLUE evaluation can be found in \nameref{protein_scores1} and \nameref{protein_scores2}. The new compute Pareto frontier highlights the benefits of multi-omic data for efficient model scaling.

Notably, results on protein evaluation tasks differed depending on whether the tokenization was per-residue/nucleotide or whether a byte-pair encoding tokenizer was used. This difference in performance is likely due to an increase in performance on per-residue tasks.

\section*{Discussion}
\subsection*{Implications, limitations, and outlook}
OmniBioTE is a series of multi-omic models (MOMs) pretrained jointly on a diverse set of nucleic acid sequences and proteomic data. We analyzed the properties of these models across a wide range of scales and tasks. We found that these models not only achieve state-of-the-art performance on single-omic tasks measured in performance-per-FLOP, but also unlock novel multi-omic tasks such as modeling protein-nucleic acid interactions by predicting the change in Gibbs free energy between a protein and nucleic acid. We also showed that as a result of this fine-tuning process, OmniBioTE learns meaningful structural information without any explicit structural training, allowing one to estimate how strongly a given protein residue or nucleotide participates in binding interactions. Although our model is able to implicitly learn some structural information, the quality of the information pales in comparison to that of models that are explicitly trained to predict the structure of protein-nucleic acid complexes. For example, AlphaFold3 \cite{alphafold3} reports an interface local distance difference test (iLDDT) score of over 55 for protein-DNA complexes, loosely meaning that, on average, AlphaFold3 predicts the interface structure between proteins and DNA within 4~\AA{} over 55\% of the time. This value is a lower bound, as iLDDT is computed as an average between distance cutoffs of 0.5, 1, 2, and 4~\AA{}, meaning that AlphaFold3 significantly outperforms the contacts predicted from the information implicitly learned by our models from $\Delta G$ fine-tuning. Multi-omic modeling of this sort is of great interest in the development of new pharmacologic therapies; many notable pharmaceutical drugs and candidate drugs that function via nucleic acid-protein interaction have already shown great promise, such as pegaptanib \cite{pegaptanib}, an RNA aptamer targeting vascular endothelial growth factor, as well as RNA sequences that target nucleolin \cite{Carvalho2019}, coagulation factors \cite{Tanaka2009, Chan2008, Riccardi2020, JilmaStohlawetz2012}, CCL2 \cite{Menne2017}, CXCL12 \cite{Giordano2024}, and hepcidin \cite{Schwoebel2013}. While our methodology does not explore aptamer design or property prediction, we believe that this methodology could be extended to aptamers with the right dataset and leave this to future research.

We found that OmniBioTE emergently learned a joint representation between protein sequences and their corresponding genes despite never explicitly being trained on a joint objective, demonstrating that training biosequence transformer models on multi-omic data can learn non-trivial representations across sequences even with a simple masked language model objective. We attribute this emergence from self-supervised pretraining as being a consequence of the efficient coding hypothesis \cite{loh2014efficient}.

A natural question is how OmniBioTE could acquire joint, modality-invariant features in the absence of explicit cross-attention between protein and nucleic-acid sequences during pretraining. We hypothesize this is due largely because the model must use the same parameters to model all sequence types. Both nucleic-acid sequences and their translated proteins can be considered as two views of the same underlying ``latent sequences'' generated from shared underlying biological factors. When such a model is trained to maximize the masked-language-modeling objective on the joint data distribution under finite capacity, the most efficient solution is to allocate some representation dimensions to the latent sequence representation and reuse them across modalities, while also learning the modality-specific information. This is the behavior predicted by efficient-coding perspectives on representation learning, in which optimization favors codes that are representative of common latent structure. Importantly, this mechanism does not require cross-attention between protein and nucleic acid tokens during pretraining; the coupling arises from the shared parameters, a common unsupervised objective over the joint data distribution, and the constraints of a finite parameter count. Analogous phenomena have been observed in multilingual language models such as mBERT \cite{mbert}, which show evidence of transfer learning and shared representations over multiple languages despite lacking any inter-language cross-attention. Our results, where a low-rank linear projector trained on only 5\% of gene–protein pairs generalizes to the remaining 95\%, are consistent with this view and suggest that OmniBioTE’s backbone already encodes a modality-invariant subspace that is approximately stable under transcription and translation.

We hypothesize that considerably richer representations could be learned if auxiliary training objectives were introduced, such as structure/property prediction, cross-attention between different modalities, or the addition of multiple sequence alignment data. Beyond additional learning objectives, we note that there has been a considerable amount of research into multi-modal vision-language modeling using novel model architectural components including cross-attention and cross-modal projectors \cite{jaegle2021perceiver, alayrac2022flamingo, radford2021learning, liu2024visual}, and that many of these approaches may be of interest in multi-modal biosequence modeling as well.

We additionally found that multi-omic pretrained models are superior or comparable at scale to identical models trained on single-omics data with identical compute budgets and smaller per-omic data budgets. Furthermore, we find that our multi-omic models set a new compute Pareto frontier across GUE and TAPE benchmarks, even before factoring in the lower amount of per-modality data each model sees during training. Despite the difference in datasets, we found no downsides to mixing in other modalities during pretraining for our biosequence foundation models in this project. In fact, our MOMs set new state-of-the-art performance numbers for several of the downstream nucleic acid tasks. Our MOMs also considerably outperformed a combination of single-omic models on the multi-omic task of binding affinity prediction, and outperformed molecular dynamics methods in conjunction with structural predictions from AlphaFold3, despite being a considerably more computationally intensive baseline. Lastly, we showed that these results robustly transfer to completely unseen and unrelated datasets by testing our models on the JASPAR dataset.

There are several notable limitations to this work that deserve special mention. Most notably, we only scratched the surface of multi-omic biosequence modeling. As noted earlier, there are many popular ways of training multi-omic sequence models, and we elected for a simple approach using a masked language modeling task. We additionally only investigate our scaling over a rough two orders of magnitude of compute and leave the training of larger models on larger datasets as future research directions that seem reasonably likely to yield performance benefits consistent with the scaling results we found in this work. Additionally, for the protein evals, the ESM models probe a higher compute budget than we were able to reach with OmniBioTE due to constraints on available compute. We hope to explore larger compute budgets in future work given the promising results at the $10^{21}$ FLOP range. Lastly, we only investigated a masked language modeling task for pretraining rather than the more popular autoregressive training framework, again leaving this approach open as a viable future research direction. 

\section*{Conclusions}
Many of biology's most significant interactions occur between proteins and nucleic acids, and we demonstrate the first large-scale attempt at building and quantifying the scaling behavior of multi-omic foundation models to specifically capture these critical molecular interactions. Our results indicate that multi-omic pretraining is a scalable and compute-efficient strategy for building unified biological foundation models with rich representations and the capacity to perform strongly on downstream multi-omic tasks. Beyond their biological significance, modeling the interactions between nucleic acids and proteins is of great pharmaceutical and clinical importance; models that can assist with the development of nucleic acids that modify the function of naturally occurring proteins would greatly accelerate pharmaceutical development. Foundational biosequence models have the promise of dramatically improving our ability to both understand and predict biology, and we hope that our work with OmniBioTE presents one of many efforts to build multi-omic models that can capture the full richness of biomolecular interactions.

\subsection*{Acknowledgments}
The authors would like to thank Michael Retchin for his insightful comments and broad literature knowledge on protein-nucleic acid interactions. The authors would like to thank Douglas Kondziolka for his feedback on the manuscript. The authors would also like to thank Vincent D'Anniballe for his helpful discussion surrounding biosequence datasets. Lastly, we would like to thank Michael Costantino and the NYU Langone High Performance Computing team for their assistance with maintaining state-of-the-art computing infrastructure necessary for this research.

\clearpage

\onecolumn

\section*{Supporting information}

\paragraph*{S1 Appendix.}
\label{S1_Appendix}
\textbf{OmniBioTE model architecture and training pipeline.} In this section we provide a detailed description of the OmniBioTE architecture and the end-to-end training pipeline used to pretrain the model on mixed nucleic acid and protein sequences. The implementation follows a standard transformer encoder architecture with minor modifications for rotary positional encodings and maximal update parameterization (µP).

\textbf{Input preprocessing and tokenization}

Each training example is a single primary sequence drawn either from GenBank (nucleic acids) or UniRef100 (proteins). Input sequences are preprocessed and tokenized as follows:

\begin{enumerate}
    \item \textbf{Sequence selection.} We sample a single raw sequence $s$ from the mixed training corpus. For multi-omic training runs, we sample from nucleic acid and protein pools.
    \item \textbf{Modality-specific tokenization.} 
    \begin{itemize}
        \item Nucleic acid sequences are tokenized with a SentencePiece tokenizer trained on GenBank entries.
        \item Protein sequences are tokenized with a SentencePiece tokenizer trained on UniRef100.
        \item Nucleic acid and protein vocabularies are strictly disjoint.
        \item For single-character models, a disjoint single-character tokenizer was used.
    \end{itemize}
    \item \textbf{Context windowing.} Tokenized sequences are split or concatenated into fixed-length segments of length 1024 for BPE models or 2048 for per-residue/nucleotide models.
\end{enumerate}

\textbf{Encoder transformer backbone}

The OmniBioTE backbone is a non-causal transformer encoder with $L$ layers and $L$ attention heads (each head has dimension $128$), and hidden width $d = 128\times L$. The main components are:

\paragraph{Token embeddings and positional encodings.}
A learned token embedding table $\mathbf{E} \in \mathbb{R}^{V \times d}$ maps each token index $x_t$ to a vector $\mathbf{e}_t = \mathbf{E}[x_t]$.

Where $V$ is the joint vocabulary size. The resulting sequence of hidden states is passed through a dropout layer and then to the transformer blocks.

\paragraph{RMS-style layer normalization.}
Each block employs an RMS normalization layer. Given an input vector $\mathbf{h} \in \mathbb{R}^{d}$, the layer computes
\[
\mu = \frac{1}{d}\sum_{i=1}^{d} h_i,\quad
\sigma = \sqrt{\frac{1}{d}\sum_{i=1}^{d}(h_i - \mu)^2 + \varepsilon},
\]
and outputs
\[
\text{RMSNorm}(\mathbf{h}) = \boldsymbol{\alpha} \odot \frac{\mathbf{h}-\mu}{\sigma} + \boldsymbol{\gamma},
\]
where $\boldsymbol{\alpha}, \boldsymbol{\gamma} \in \mathbb{R}^d$ are learned scale and shift parameters and $\varepsilon$ is a small constant (here $10^{-5}$).

\paragraph{Multi-head self-attention with RoPE\cite{su2023roformerenhancedtransformerrotary}.}
For each block, the self-attention layer maps an input sequence $\mathbf{X} \in \mathbb{R}^{T \times d}$ to an output $\mathbf{Y} \in \mathbb{R}^{T \times d}$:
\begin{enumerate}
    \item A single linear projection produces concatenated queries, keys, and values:
    \[
    [\mathbf{Q}, \mathbf{K}, \mathbf{V}] = \mathbf{X} \mathbf{W}_{\text{qkv}} \quad
    \mathbf{W}_{\text{qkv}} \in \mathbb{R}^{d \times 3d}.
    \]
    \item The resulting matrices are reshaped into $H$ attention heads of dimension $d_h=128$:
    \[
    \mathbf{Q}, \mathbf{K}, \mathbf{V} \in \mathbb{R}^{H \times T \times d_h}.
    \]
    \item Rotary positional embeddings are applied to $\mathbf{Q}$ and $\mathbf{K}$. For each head and time index $t$, we interpret the last dimension as complex pairs and multiply by a precomputed complex sinusoid $\omega_{t} \in \mathbb{C}^{d_h/2}$:
    \[
    \tilde{\mathbf{Q}}_{t}, \tilde{\mathbf{K}}_{t} = \text{RoPE}(\mathbf{Q}_{t}, \mathbf{K}_{t}; \omega_{t}).
    \]
    \item Full non-causal attention is computed using PyTorch's fused scaled dot-product attention. The attention output per head is
    \[
    \mathbf{A} = \text{Attention}(\tilde{\mathbf{Q}}, \tilde{\mathbf{K}}, \mathbf{V}),
    \]
    where the attention logit scaling factor is set to $\alpha = 8 / d$ to match µP scaling.
    \item The head outputs are concatenated and projected back to the model dimension:
    \[
    \mathbf{Y} = \text{Dropout}\left(\text{Concat}_{h=1}^{H}(\mathbf{A}^{(h)}) \mathbf{W}_{\text{o}}\right),
    \]
    with $\mathbf{W}_{\text{o}} \in \mathbb{R}^{d \times d}$.
\end{enumerate}

\paragraph{SwiGLU MLP.}
Following the attention layer, each block applies a SwiGLU feed-forward network:
\begin{enumerate}
    \item A linear layer expands the hidden dimension by a factor of 4:
    \[
    \mathbf{U} = \mathbf{X}\mathbf{W}_{\text{ff}},\quad \mathbf{W}_{\text{ff}} \in \mathbb{R}^{d \times 4d}.
    \]
    \item The expanded representation is split into a value component $\mathbf{A}$ and a gating component $\mathbf{B}$ along the last dimension and passed through a SwiGLU nonlinearity.
    \item A projection back to dimension $d$ produces the MLP output, which is then passed through dropout:
    \[
    \mathbf{Y}_{\text{MLP}} = \text{Dropout}(\mathbf{Z}\mathbf{W}_{\text{proj}}).
    \]
\end{enumerate}

The MLP uses the same µP-compatible initialization as the attention projections: weights are initialized with standard deviation $\sigma_\text{param} = 1/\sqrt{d}$ and the output projection is additionally scaled by $1/\sqrt{2}$ because its input width is halved by SwiGLU.

\paragraph{Residual block structure.}
Each transformer block combines these components via pre-normalization and residual connections:
\begin{align*}
\mathbf{h}_0 &= \mathbf{x},\\
\mathbf{h}_1 &= \mathbf{h}_0 + \text{SelfAttention}(\text{RMSNorm}(\mathbf{h}_0)),\\
\mathbf{h}_2 &= \mathbf{h}_1 + \text{MLP}(\text{RMSNorm}(\mathbf{h}_1)).
\end{align*}
After $L$ such blocks, a final RMS normalization layer produces the encoder output $\mathbf{Z} \in \mathbb{R}^{T \times d}$.

\textbf{Output head and masked language modeling objective}

For pretraining, OmniBioTE is trained with a masked language modeling (MLM) objective over both modalities:

\begin{enumerate}
    \item \textbf{Masking strategy.} For each sequence, we independently mask each token with probability $p_\text{mask} = 0.15$. Masked positions are replaced by a dedicated MASK token. The original unmasked token indices are retained as the prediction targets.
    \item \textbf{Projection head.} The final hidden states $\mathbf{Z} \in \mathbb{R}^{T \times d}$ are projected to logits over the vocabulary with a single linear layer whose weights are not shared with the input embedding matrix:
    \[
    \mathbf{L} = \mathbf{Z} \mathbf{W}_{\text{lm}}, \quad \mathbf{W}_{\text{lm}} \in \mathbb{R}^{d \times V}.
    \]
    To maintain µP scaling, the logits are multiplied by a global factor
    \[
    \mathbf{L}' = \left(\frac{32}{d}\right)\mathbf{L},
    \]
    which allows learning rates to be specified in a width-invariant fashion.
    \item \textbf{Loss computation.} Cross-entropy loss is computed only on masked positions. Let $\mathcal{M}$ denote the set of masked indices in the batch. The MLM loss is
    \[
    \mathcal{L}_\text{MLM} = -\frac{1}{|\mathcal{M}|} \sum_{(b,t) \in \mathcal{M}} \log p\left(x_{b,t} \mid x_{b,\neg t}; \theta\right),
    \]
    where $p(\cdot)$ is the softmax over $\mathbf{L}'$ and $x_{b,\neg  t}$ denotes all tokens in the sequence excluding position $t$.
\end{enumerate}

\textbf{Training}
\begin{itemize}
    \item \textbf{Batch sizing.} The global batch size $B_\text{global}$ is chosen so that the token throughput approaches the memory/computation limit of the cluster. In practice we used global batch sizes in the range of 786{,}432–1{,}048{,}576 tokens per step.
    \item \textbf{Optimizer.} We use AdamW\cite{loshchilov2019decoupledweightdecayregularization} and µP learning rates:
    \begin{itemize}
        \item Embedding and unembedding layers, as well as all bias and normalization parameters, are assigned a \emph{fixed} learning rate.
        \item All remaining weights are assigned a learning rate that scales with $32/d$, and their weight decay is scaled inversely so that the effective decrease per optimization step is width-invariant.
    \end{itemize}
    \item \textbf{Scheduler.} A OneCycleLR\cite{smith2018superconvergencefasttrainingneural} schedule is applied over the entire training run, with an initial warmup period of 1 billion tokens and decay to the original learning rate multiplied by $10^{-5}$.
\end{itemize}

This architecture and training setup is identical for all OmniBioTE sizes, with only $(L, H, d)$ and the tokenization (BPE vs per-residue/nucleotide) varying between model families.

\paragraph*{S2 Appendix.}
\label{S2_Appendix}
\textbf{Predicting binding interactions between proteins and nucleic acids with AlphaFold3 and molecular dynamics.} We developed and assessed a pipeline for predicting the interaction energy between proteins and nucleic acids by combining AlphaFold3 (AF3) \cite{alphafold3} with molecular dynamics (MD) simulations performed in OpenMM version 8.1.2 \cite{eastman2023openmm}. 
Our pipeline is available at the link \url{https://github.com/hockyg/af3_protein_nucleic_md_pipeline}.
We applied this pipeline to as many targets as possible from the ProNAB database studied in Fig~\ref{fig:multiomics}. Ultimately, we were able to generate structure predictions and perform MD simulations on 599 protein/nucleic acid pairs.

Given that we wanted to compare our ability to predict binding energies from sequence directly, this required us to generate bound conformations via a machine learning approach that treats both proteins and nucleic acids, and so for that we selected one of the only available options, AF3 \cite{alphafold3}. Due to the large size of the systems and the need to rapidly evaluate interactions, we were forced to use an \textit{implicit solvent} approach. This need was exacerbated by the fact that AF3 predicted structures often have large regions with low confidence scores that are non-compact, resulting in simulation boxes that would be intractable if filled with water (i.e. millions of atoms including solvent and ions). Using implicit solvent, the MD simulations executed for our targets ranged in size from 1195 atoms to 60832 atoms, with an average size of approximately 7653.

To approximately compute the binding energy between a protein and nucleic acid in tractable computational time, we adopted a protocol similar to the so-called MM/GBSA approach \cite{wang2019end}. 
To compute the binding free energy of a complex, we need to compute \begin{equation}
    \Delta G = G_\mathrm{AB}-G_\mathrm{A}-G_\mathrm{B},
\end{equation}

where $A$ and $B$ are the separate components and the free energies on each side are averaged over a conformational ensemble. 
$\Delta G$ has contributions that come from the (1) direct interaction energy between the molecules, (2) the change in solvation free energy due to the difference in buried surface area, (3) and the change in configurational entropy of both parts upon binding. 
When using simulations with implicit solvent, effects 1 and 2 are taken into effect if we simply calculate the MD energy. The third effect due to overall changes in the conformations of the bound and unbound $A$ and $B$ molecules is not possible to calculate in a single simulation and requires extensive calculations beyond the scope of this work.
However, conveniently, we expect that for calculations of $\Delta \Delta G$ of mutation, this term cancels out.
Below, we will therefore run short MD simulations and compute the energy of the complex as well as for separate components in order to see whether $\Delta G^\mathrm{experiment}$ can be predicted.
 \textit{We emphasize that we do not expect this to work in general \cite{roux2024editorial}, and we are performing these calculations to set a baseline for our ML predictions given in the main text.}

To go from sequence to energy prediction, we start by converting entries in the ProNAB database \cite{10.1093/nar/gkab848} into YAML files suitable for AF3 predictions. 
This consists of specifying a protein chain and a nucleic acid chain (or chains in the case of a double stranded sequence).
We also added 1 Mg$^{2+}$ ion per nucleotide in case explicit divalent cations were needed for solvated MD simulations in the future. These divalent ions were removed for implicit solvent simulations performed next.

The topology of the system was built from the AF3 output CIF file using PDBFixer in OpenMM \cite{eastman2023openmm}. 
The forcefield used was Amber14 \cite{maier2015ff14sb} with the GB2 implicit solvent model \cite{nguyen2013improved}.
After minimizing the energy, the velocities were randomized and the system was equilibrated at $T=300$K by running 100 ps of MD, before running 10 ns of MD using a 4 fs timestep and the \texttt{LangevinMiddle} integrator \cite{leimkuhler2013robust,zhang2019unified} with hydrogen mass repartitioning to a mass of 3 amu and a drag coefficient of 1 ps$^{-1}$. 
The energies of the full complex, and the protein and nucleic acid separately were averaged over the final 5 ns of MD simulation to produce the $\Delta E_\mathrm{pred}=E_{AB}-E_A-E_B$ values in \nameref{S1_Fig}.

\paragraph*{S1 Fig.}
\label{S1_Fig}
\textbf{Comparison of experimental binding energies with metrics predicted by AlphaFold3 and molecular dynamics}. \textbf{A} Comparison of energy differences between bound and separated states using MD as described in the text with experimentally measured values. Squares denote proteins bound to DNA while circles are proteins bound to RNA. Color indicates a different protein. (\textbf{B}) Relative predicted free energy difference between binding a native and mutant nucleic acid sequence. (\textbf{C}) Comparison of experimental binding free energies with AF3 pTM scores (0-1 with 1 being a high confidence in binding prediction).

\begin{figure}[t]
    \centering
    \includegraphics[width=1.0\textwidth]{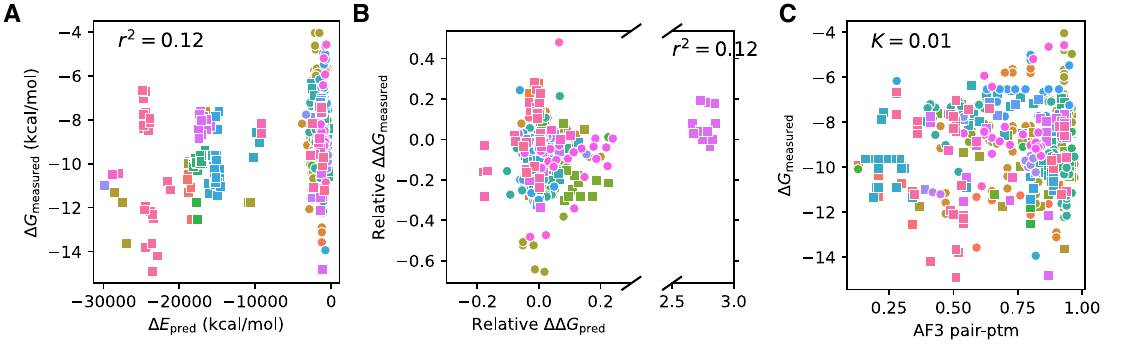}
    \caption{(\textbf{A}) Comparison of energy differences between bound and separated states using MD as described in the text with experimentally measured values. Squares denote proteins bound to DNA while circles are proteins bound to RNA. Color indicates a different protein. (\textbf{B}) Relative predicted free energy difference between binding a native and mutant nucleic acid sequence. (\textbf{C}) Comparison of experimental binding free energies with AF3 pTM scores (0-1 with 1 being a high confidence in binding prediction). }
    \label{fig:MD}
\end{figure}

\nameref{S1_Fig} shows a scatter plot comparing $\Delta E_\mathrm{pred}$ with measured binding free energies for those complexes. Multiple values are given for the same protein when different nucleic acid sequences were given in the database for mutational studies. 
As can be seen, both the order of magnitude is greatly different, and also there is little to no correlation as measured by the square of the Pearson correlation, $r^2$.

In \nameref{S1_Fig} B we show the results of computing a predicted $\Delta \Delta G_\mathrm{pred} \approx \Delta E_\mathrm{pred}^\mathrm{mutant}-\Delta E_\mathrm{pred}^\mathrm{native}$ to those from experiment.
The absolute values of $\Delta \Delta G_\mathrm{pred}$ were much larger than for experiment again, so here we show the relative difference (i.e. scaled by the native binding free energy or energy value). Again here there is no correlation. There is also one 
 target (p04390) which is an outlier, but there is nothing obvious suggesting why this protein shows such a large relative error compared to the others.
 There is also little to no correlation when removing this outlier. 

 We check whether the pTM confidence scores reported by AF3 \cite{alphafold3} are correlated with the binding affinity for these complexes. This metric also has no correlation, as measured by the Spearman rank order correlation coefficient, $K$ (\nameref{S1_Fig} C).

 Finally, we can also consider the time required to compute these results. Simulations that we attempted ranged from 5 to 912 ns/day on a single GPU. 
 For those simulations that completed the full 10 ns of MD, times ranged from approximately 0.01 to 0.2 hours, for a total of approximately 13.4 GPU-hours.
While this already far exceeds the inference time from our ML model, this was not the time consuming part of the calculation. 
AF3 calculations can be split into two parts, one which involves a multiple sequence alignment (MSA), and then the actual inference \cite{alphafold3}. While the inference step is relatively fast, the MSA step is slow and CPU bound, and was the longest part of the calculation.
For these calculations, the average inference took 0.9$\pm$0.4 minutes on an A100 GPU, while the average MSA computation time took 47.5 minutes and ranged from 14 to 283 minutes on 16 CPUs, for a total of over 6700 CPU-hours.

\paragraph*{S1 Table.}
\label{dataset_statistics}
\textbf{Training data statistics across all sequence types.}

\begin{table}[!h]
\centering
\begin{tabular}{|l|r|r|r|}
\hline
\textbf{Sequence Type} & \makecell{\textbf{Average Length} \\ (bp/residues)} & \makecell{\textbf{Minimum Length} \\ (bp/residues)} & \makecell{\textbf{Maximum Length} \\ (bp/residues)} \\ \hline
\Xhline{2\arrayrulewidth}
DNA          & 16941.82  $\pm$ 1421192.40 & 6    & 363684565 \\ \hline
mRNA         & 624.75 $\pm$ 539.25       & 6    & 84308 \\ \hline
RNA          & 10027.20 $\pm$ 39070.07   & 6    & 167463040 \\ \hline
cRNA         & 1769.64 $\pm$ 1945.04     & 35   & 157276 \\ \hline
rRNA         & 482.19 $\pm$ 266.62       & 24   & 7097 \\ \hline
ss-RNA       & 3087.72 $\pm$ 4797.85     & 14   & 35911 \\ \hline
ss-DNA       & 1637.48 $\pm$ 1253.90     & 17   & 34395 \\ \hline
ds-RNA       & 2075.30 $\pm$ 2197.43     & 48   & 31081 \\ \hline
tRNA         & 249.23 $\pm$ 349.17       & 20   & 1208 \\ \hline
ds-cRNA      & 657.16 $\pm$ 970.66       & 127  & 15341 \\ \hline
ms-DNA       & 15492.76 $\pm$ 19181.55   & 84   & 45513 \\ \hline
ds-mRNA      & 1695.75 $\pm$ 547.50      & 1114 & 2414 \\ \hline
ms-RNA       & 606.50 $\pm$ 28.50        & 578  & 635 \\ \hline
ds-rRNA      & 580.00 $\pm$ 0.00         & 580  & 580 \\ \hline
peptide      & 388.19 $\pm$ 379.80       & 5    & 45359 \\ \hline
\end{tabular}
\end{table}

\clearpage

\paragraph*{S2 Table.}
\label{pdb_contact_f1} 
\textbf{Mean F1 scores for predicted contact maps at distance thresholds of 4 \AA, 6 \AA, and 8 \AA.}

\begin{table}[!h]
\centering
\begin{tabular}{|l|c|c|c|}
\hline
\multirow{2}{*}{Model} & \multicolumn{3}{c|}{Mean F1 score $\pm$ standard deviation \; (N = 10)} \\ \cline{2-4}
 & 4 \AA & 6 \AA & 8 \AA \\ \hline
\multicolumn{4}{|c|}{\textbf{OmniBioTE}} \\ \hline
Small  & 0.6821 $\pm$ 0.0760 & 0.7428 $\pm$ 0.0752 & 0.8134 $\pm$ 0.0772 \\ \hline
Medium & 0.6951 $\pm$ 0.0595 & 0.7585 $\pm$ 0.0584 & 0.8254 $\pm$ 0.0552 \\ \hline
Large  & 0.6934 $\pm$ 0.0609 & 0.7573 $\pm$ 0.0588 & 0.8230 $\pm$ 0.0562 \\ \hline
XL     & 0.6962 $\pm$ 0.0546 & 0.7574 $\pm$ 0.0520 & 0.8264 $\pm$ 0.0501 \\ \hline
\multicolumn{4}{|c|}{\textbf{OmniBioTE (per-residue/nucleotide)}} \\ \hline
Small  & 0.8287 $\pm$ 0.1513 & 0.8675 $\pm$ 0.1410 & 0.8987 $\pm$ 0.1148 \\ \hline
Medium & 0.8773 $\pm$ 0.0789 & 0.9077 $\pm$ 0.0780 & 0.9286 $\pm$ 0.0678 \\ \hline
Large  & 0.8789 $\pm$ 0.0792 & 0.9105 $\pm$ 0.0747 & 0.9310 $\pm$ 0.0651 \\ \hline
XL     & 0.8796 $\pm$ 0.0740 & 0.9116 $\pm$ 0.0722 & 0.9312 $\pm$ 0.0620 \\ \hline
\multicolumn{4}{|c|}{\textbf{LucaOne}} \\ \hline
LucaOne             & 0.7680 $\pm$ 0.1869 & 0.8270 $\pm$ 0.1546 & 0.8713 $\pm$ 0.1231 \\ \hline
\end{tabular}
\end{table}

\clearpage

\paragraph*{S3 Table.}
\label{pronab_results}
\textbf{OmniBioTE performance across all 10-folds of the Pronab mutation benchmark as measured in Pearson correlation coefficient (PCC) and mean absolute error (MAE).}

\begin{table}[!h]
\centering
\begin{tabular}{|l|c|c|}
\hline
Model & $\Delta G$ PCC & $\Delta G$ MAE \\ \hline
\Xhline{2\arrayrulewidth}
OmniBioTE-small & 0.255 $\pm$ 0.095 & 1.68 $\pm$ 0.22 \\ \hline
OmniBioTE-medium & 0.372 $\pm$ 0.073 & 1.57 $\pm$ 0.20 \\ \hline
OmniBioTE-large & 0.346 $\pm$ 0.079 & 1.60 $\pm$ 0.21 \\ \hline
OmniBioTE-XL & \textbf{0.407 $\pm$ 0.098} & \textbf{1.56 $\pm$ 0.23} \\ \hline
\Xhline{2\arrayrulewidth}
OmniBioTE-small (per-nucleotide/residue) & 0.227 $\pm$ 0.122 & 1.72 $\pm$ 0.20 \\ \hline
OmniBioTE-medium (per-nucleotide/residue) & 0.335 $\pm$ 0.065 & 1.63 $\pm$ 0.22 \\ \hline
OmniBioTE-large (per-nucleotide/residue) & 0.299 $\pm$ 0.127 & 1.63 $\pm$ 0.29 \\ \hline
OmniBioTE-XL (per-nucleotide/residue) & 0.378 $\pm$ 0.088 & 1.56 $\pm$ 0.25 \\ \hline
\Xhline{2\arrayrulewidth}
Nuc+ProtBioTE-small & 0.071 $\pm$ 0.112 & 1.71 $\pm$ 0.24 \\ \hline
Nuc+ProtBioTE-medium & 0.081 $\pm$ 0.116 & 1.83 $\pm$ 0.32 \\ \hline
Nuc+ProtBioTE-large & 0.039 $\pm$ 0.134 & 1.83 $\pm$ 0.29 \\ \hline
Nuc+ProtBioTE-XL & 0.077 $\pm$ 0.140 & 1.90 $\pm$ 0.29 \\ \hline
\Xhline{2\arrayrulewidth}
LucaOne & 0.200 $\pm$ 0.16 & 2.429 $\pm$ 0.293 \\ \hline
DeePNAP & 0.100 $\pm$ 0.111 & 2.442 $\pm$ 0.821 \\ \hline
AlphaFold3 + simulation & 0.332 & -- \\ \hline
\end{tabular}
\end{table}

\clearpage

\paragraph*{S4 Table.}
\label{epigenetics1}
\textbf{GUE Results (Epigenetics): Histone Modification Benchmarks (Part 1). Values represent the Matthews correlation coefficient of the predictions.}

\begin{table}[!h]
\centering
\resizebox{\textwidth}{!}{%
\begin{tabular}{|l|r|r|r|r|r|r|}
\hline
\multicolumn{7}{|c|}{OmniBioTE} \\ \hline
\Xhline{2\arrayrulewidth}
Model & H3 & H3K14ac & H3K36me3 & H3K4me1 & H3K4me2 & H3K4me3 \\ \hline
OmniBioTE-small  & 77.01 & 58.23 & 59.42 & 51.83 & 33.62 & 37.89 \\ \hline
OmniBioTE-medium & 79.75 & 66.40 & 68.01 & 60.03 & 49.56 & 55.02 \\ \hline
OmniBioTE-large  & 80.64 & 67.31 & 69.48 & 59.04 & 46.64 & 55.33 \\ \hline
OmniBioTE-XL     & 82.11 & 67.34 & 70.22 & 58.14 & 52.45 & 57.43 \\ \hline
\Xhline{2\arrayrulewidth}
\multicolumn{7}{|c|}{OmniBioTE (per-nucleotide)} \\ \hline
OmniBioTE-small (per-nucleotide)  & 77.85 & 53.96 & 60.93 & 54.67 & 30.32 & 30.32 \\ \hline
OmniBioTE-medium (per-nucleotide) & 80.76 & 56.94 & 63.58 & 54.69 & 32.52 & 45.27 \\ \hline
OmniBioTE-large (per-nucleotide)  & 82.64 & 71.52 & 69.89 & 62.42 & 57.56 & 59.33 \\ \hline
OmniBioTE-XL (per-nucleotide)     & 80.47 & 59.27 & 66.20 & 54.59 & 45.71 & 47.16 \\ \hline
\Xhline{2\arrayrulewidth}
\multicolumn{7}{|c|}{NucBioTE} \\ \hline
NucBioTE-small  & 76.93 & 53.83 & 56.46 & 46.81 & 36.11 & 40.34 \\ \hline
NucBioTE-medium & 75.44 & 49.76 & 59.04 & 38.98 & 27.55 & 35.10 \\ \hline
NucBioTE-large  & 76.51 & 53.51 & 55.45 & 47.05 & 32.68 & 40.71 \\ \hline
NucBioTE-XL     & 80.81 & 66.91 & 66.44 & 55.26 & 47.20 & 57.04 \\ \hline
\Xhline{2\arrayrulewidth}
\multicolumn{7}{|c|}{Baselines} \\ \hline
HyenaDNA \cite{hyenadna}                         & 67.17 & 31.98 & 48.27 & 35.83 & 25.81 & 23.15 \\ \hline
NT-2500M-multi \cite{dalla2023nucleotide}      & 78.77 & 56.20 & 61.99 & 55.30 & 36.49 & 40.34 \\ \hline
DNABERT-2 \cite{zhou2024dnabert2efficientfoundationmodel} & 78.27 & 52.57 & 56.88 & 50.52 & 31.13 & 36.27 \\ \hline
RandomMask \cite{randommask}                     & 77.62 & 65.07 & 63.68 & 54.47 & 53.88 & 62.19 \\ \hline
LucaOne                                         & 72.28 & 44.61 & 46.72 & 42.33 & 28.79 & 25.96 \\ \hline
\end{tabular}
}
\end{table}

\clearpage

\paragraph*{S5 Table.}
\label{epigenetics2}
\textbf{GUE Results (Epigenetics): Histone Modification Benchmarks (Part 2). Values represent the Matthews correlation coefficient of the predictions.}

\begin{table}[!h]
\centering
\begin{tabular}{|l|r|r|r|r|}
\hline
Model & H3K79me3 & H3K9ac & H4 & H4ac \\ \hline
\Xhline{2\arrayrulewidth}
\multicolumn{5}{|c|}{OmniBioTE} \\ \hline
OmniBioTE-small  & 63.48 & 61.94 & 79.70 & 47.12 \\ \hline
OmniBioTE-medium & 72.99 & 68.79 & 82.50 & 62.57 \\ \hline
OmniBioTE-large  & 72.57 & 67.99 & 82.50 & 63.62 \\ \hline
OmniBioTE-XL     & 73.35 & 66.75 & 81.55 & 63.71 \\ \hline
\Xhline{2\arrayrulewidth}
\multicolumn{5}{|c|}{OmniBioTE (per-nucleotide)} \\ \hline
OmniBioTE-small (per-nucleotide)  & 63.14 & 57.13 & 81.94 & 46.86 \\ \hline
OmniBioTE-medium (per-nucleotide) & 67.62 & 59.17 & 82.54 & 52.22 \\ \hline
OmniBioTE-large (per-nucleotide)  & 73.69 & 67.91 & 83.48 & 65.86 \\ \hline
OmniBioTE-XL (per-nucleotide)     & 73.07 & 60.89 & 80.90 & 58.18 \\ \hline
\Xhline{2\arrayrulewidth}
\multicolumn{5}{|c|}{NucBioTE} \\ \hline
NucBioTE-small  & 63.17 & 54.31 & 78.69 & 52.12 \\ \hline
NucBioTE-medium & 62.50 & 51.78 & 79.86 & 38.15 \\ \hline
NucBioTE-large  & 66.78 & 58.41 & 80.84 & 50.19 \\ \hline
NucBioTE-XL     & 72.73 & 65.95 & 82.65 & 62.41 \\ \hline
\Xhline{2\arrayrulewidth}
\multicolumn{5}{|c|}{Baselines} \\ \hline
HyenaDNA \cite{hyenadna}                         & 54.09 & 50.84 & 73.69 & 38.44 \\ \hline
NT-2500M-multi \cite{dalla2023nucleotide}      & 64.70 & 56.01 & 81.67 & 49.13 \\ \hline
DNABERT-2 \cite{zhou2024dnabert2efficientfoundationmodel} & 67.39 & 55.63 & 80.71 & 50.43 \\ \hline
RandomMask \cite{randommask}                     & 72.67 & 65.02 & 79.44 & 64.22 \\ \hline
LucaOne                                         & 59.69 & 50.82 & 76.24 & 36.70 \\ \hline
\end{tabular}
\end{table}

\clearpage

\paragraph*{S6 Table.}
\label{tf_covid}
\textbf{GUE Results: Human Transcription Factors and COVID. Values represent the Matthews correlation coefficient of the predictions, with the exception of the COVID variant prediction task which uses F1-score.}

\begin{table}[!h]
\centering
\begin{tabular}{|l|r|r|r|r|r|r|}
\hline
\multirow{2}{*}{Model} & \multicolumn{5}{c|}{Human Transcription Factors} & Covid \\ \cline{2-7}
 & 0 & 1 & 2 & 3 & 4 &  \\ \hline
 \Xhline{2\arrayrulewidth}
\multicolumn{7}{|c|}{OmniBioTE} \\ \hline
OmniBioTE-small  & 65.67 & 70.07 & 56.43 & 46.36 & 65.81 & 67.93 \\ \hline
OmniBioTE-medium & 62.37 & 72.04 & 59.63 & 47.22 & 76.02 & 69.38 \\ \hline
OmniBioTE-large  & 62.53 & 72.08 & 60.40 & 51.94 & 75.76 & 69.26 \\ \hline
OmniBioTE-XL     & 64.82 & 69.95 & 63.75 & 55.44 & 75.65 & 68.77 \\ \hline
\Xhline{2\arrayrulewidth}
\multicolumn{7}{|c|}{OmniBioTE (per-nucleotide)} \\ \hline
OmniBioTE-small (per-nucleotide)  & 64.80 & 70.83 & 53.22 & 45.29 & 73.00 & 57.30 \\ \hline
OmniBioTE-medium (per-nucleotide) & 66.86 & 69.08 & 69.12 & 51.34 & 77.69 & 73.50 \\ \hline
OmniBioTE-large (per-nucleotide)  & 65.77 & 70.46 & 67.49 & 51.62 & 77.74 & 76.55 \\ \hline
OmniBioTE-XL (per-nucleotide)     & 66.50 & 67.82 & 62.95 & 53.32 & 76.02 & 74.11 \\ \hline
\Xhline{2\arrayrulewidth}
\multicolumn{7}{|c|}{NucBioTE} \\ \hline
NucBioTE-small  & 65.50 & 69.92 & 53.82 & 38.98 & 74.00 & 66.02 \\ \hline
NucBioTE-medium & 64.19 & 66.98 & 53.50 & 50.28 & 73.03 & 59.66 \\ \hline
NucBioTE-large  & 63.50 & 65.24 & 56.67 & 41.90 & 69.28 & 67.01 \\ \hline
NucBioTE-XL     & 64.78 & 68.50 & 59.15 & 43.18 & 76.83 & 67.82 \\ \hline
\Xhline{2\arrayrulewidth}
\multicolumn{7}{|c|}{Baselines} \\ \hline
HyenaDNA \cite{hyenadna}                    & 62.30 & 67.86 & 46.85 & 41.78 & 61.23 & 23.27 \\ \hline
NT-2500M-multi \cite{dalla2023nucleotide} & 66.64 & 70.28 & 58.72 & 51.65 & 69.34 & 73.04 \\ \hline
DNABERT-2 \cite{zhou2024dnabert2efficientfoundationmodel} & 71.99 & 76.06 & 66.52 & 58.54 & 77.43 & 71.02 \\ \hline
RandomMask \cite{randommask}                & 67.13 & 72.55 & 71.64 & 60.14 & 77.20 & -- \\ \hline
LucaOne                                    & 66.84 & 69.00 & 57.23 & 41.25 & 67.83 & 38.92 \\ \hline
\end{tabular}
\end{table}

\clearpage

\paragraph*{S7 Table.}
\label{mouse_tf}
\textbf{GUE Results: Mouse Transcription Factors. Values represent the Matthews correlation coefficient of the predictions.}

\begin{table}[!h]
\centering
\begin{tabular}{|l|r|r|r|r|r|}
\hline
\multirow{2}{*}{Model} & \multicolumn{5}{c|}{Mouse Transcription Factors} \\ \cline{2-6}
 & 0 & 1 & 2 & 3 & 4 \\ \hline
 \Xhline{2\arrayrulewidth}
\multicolumn{6}{|c|}{OmniBioTE} \\ \hline
OmniBioTE-small  & 46.67 & 82.67 & 81.71 & 68.29 & 43.07 \\ \hline
OmniBioTE-medium & 56.42 & 84.94 & 79.88 & 70.78 & 47.96 \\ \hline
OmniBioTE-large  & 57.38 & 84.60 & 76.33 & 78.01 & 49.70 \\ \hline
OmniBioTE-XL     & 60.50 & 85.01 & 83.61 & 83.26 & 52.01 \\ \hline
\Xhline{2\arrayrulewidth}
\multicolumn{6}{|c|}{OmniBioTE (per-nucleotide)} \\ \hline
OmniBioTE-small (per-nucleotide)  & 37.79 & 82.00 & 75.62 & 71.90 & 39.93 \\ \hline
OmniBioTE-medium (per-nucleotide) & 64.07 & 85.47 & 85.39 & 80.82 & 52.33 \\ \hline
OmniBioTE-large (per-nucleotide)  & 63.83 & 84.86 & 83.55 & 84.24 & 51.43 \\ \hline
OmniBioTE-XL (per-nucleotide)     & 63.95 & 85.60 & 81.10 & 87.52 & 53.05 \\ \hline
\Xhline{2\arrayrulewidth}
\multicolumn{6}{|c|}{NucBioTE} \\ \hline
NucBioTE-small  & 48.92 & 82.95 & 73.22 & 70.83 & 41.58 \\ \hline
NucBioTE-medium & 52.62 & 82.63 & 77.76 & 69.22 & 40.76 \\ \hline
NucBioTE-large  & 48.34 & 81.23 & 72.00 & 69.91 & 37.15 \\ \hline
NucBioTE-XL     & 53.11 & 83.38 & 73.85 & 63.73 & 48.65 \\ \hline
\Xhline{2\arrayrulewidth}
\multicolumn{6}{|c|}{Baselines} \\ \hline
HyenaDNA \cite{hyenadna}                    & 35.62 & 80.50 & 65.34 & 54.20 & 19.17 \\ \hline
NT-2500M-multi \cite{dalla2023nucleotide} & 63.31 & 83.76 & 71.52 & 69.44 & 47.07 \\ \hline
DNABERT-2 \cite{zhou2024dnabert2efficientfoundationmodel} & 56.76 & 84.77 & 79.32 & 66.47 & 52.66 \\ \hline
RandomMask \cite{randommask}                & 55.61 & 82.72 & 77.61 & 74.06 & 49.81 \\ \hline
LucaOne                                    & 52.33 & 82.57 & 73.44 & 57.11 & 45.17 \\ \hline
\end{tabular}
\end{table}

\clearpage

\paragraph*{S8 Table.}
\label{promoter_detection}
\textbf{Promoter Detection performance across all promoters (All) and promoter subtypes (No TATA, TATA).}
\begin{table}[!h]
\centering
\begin{tabular}{|l|r|r|r|}
\hline
\multirow{2}{*}{Model} & All & \multicolumn{2}{c|}{Promoter Type} \\ \cline{3-4}
 &  & No TATA & TATA \\ \hline
\Xhline{2\arrayrulewidth}
\multicolumn{4}{|c|}{OmniBioTE} \\ \hline
OmniBioTE-XL   & 88.99 & 94.05 & 68.38 \\ \hline
OmniBioTE-L    & 89.43 & 93.48 & 65.71 \\ \hline
OmniBioTE-M    & 88.59 & 94.17 & 68.99 \\ \hline
OmniBioTE-S    & 87.07 & 92.17 & 63.45 \\ \hline
\Xhline{2\arrayrulewidth}
\multicolumn{4}{|c|}{OmniBioTE (per-nucleotide)} \\ \hline
OmniBioTE-XL (per-nucleotide) & 93.39 & 95.25 & 85.63 \\ \hline
OmniBioTE-L  (per-nucleotide) & 94.80 & 83.37 & 70.88 \\ \hline
OmniBioTE-M  (per-nucleotide) & 92.91 & 94.69 & 81.73 \\ \hline
OmniBioTE-S  (per-nucleotide) & 92.41 & 93.48 & 86.99 \\ \hline
\Xhline{2\arrayrulewidth}
\multicolumn{4}{|c|}{NucBioTE} \\ \hline
NucBioTE-XL   & 89.50 & 93.78 & 68.20 \\ \hline
NucBioTE-L    & 85.37 & 90.43 & 65.39 \\ \hline
NucBioTE-M    & 83.99 & 91.60 & 65.87 \\ \hline
NucBioTE-S    & 86.56 & 92.39 & 65.39 \\ \hline
\Xhline{2\arrayrulewidth}
\multicolumn{4}{|c|}{Baselines} \\ \hline
HyenaDNA \cite{hyenadna}                    & 47.38 & 52.24 &  5.34 \\ \hline
NT-2500M-multi \cite{dalla2023nucleotide}   & 91.01 & 94.00 & 79.43 \\ \hline
DNABERT-2 \cite{zhou2024dnabert2efficientfoundationmodel} & 86.77 & 94.27 & 71.59 \\ \hline
RandomMask \cite{randommask}                & 92.74 & 93.40 & 84.03 \\ \hline
LucaOne                                     & 84.50 & 91.86 & 62.12 \\ \hline
\end{tabular}
\end{table}

\clearpage

\paragraph*{S9 Table.}
\label{core_promoter}
\textbf{Core Promoter evaluation: performance across all promoters (All) and promoter subtypes (No TATA, TATA).}

\begin{table}[!h]
\centering
\begin{tabular}{|l|r|r|r|}
\hline
\multirow{2}{*}{Model} & All & \multicolumn{2}{c|}{Promoter Type} \\ \cline{3-4}
 &  & No TATA & TATA \\ \hline
\Xhline{2\arrayrulewidth}
\multicolumn{4}{|c|}{OmniBioTE} \\ \hline
OmniBioTE-XL   & 64.49 & 66.09 & 73.38 \\ \hline
OmniBioTE-L    & 63.99 & 65.09 & 73.29 \\ \hline
OmniBioTE-M    & 63.72 & 66.96 & 78.41 \\ \hline
OmniBioTE-S    & 63.53 & 65.93 & 73.01 \\ \hline
\Xhline{2\arrayrulewidth}
\multicolumn{4}{|c|}{OmniBioTE (per-nucleotide)} \\ \hline
OmniBioTE-XL (per-nucleotide) & 70.34 & 71.33 & 79.37 \\ \hline
OmniBioTE-L  (per-nucleotide) & 70.88 & 71.78 & 84.96 \\ \hline
OmniBioTE-M  (per-nucleotide) & 70.75 & 70.57 & 83.38 \\ \hline
OmniBioTE-S  (per-nucleotide) & 71.83 & 70.88 & 82.43 \\ \hline
\Xhline{2\arrayrulewidth}
\multicolumn{4}{|c|}{NucBioTE} \\ \hline
NucBioTE-XL    & 63.11 & 65.33 & 62.51 \\ \hline
NucBioTE-L     & 59.73 & 63.78 & 71.35 \\ \hline
NucBioTE-M     & 63.41 & 64.61 & 71.11 \\ \hline
NucBioTE-S     & 69.21 & 65.76 & 74.32 \\ \hline
\Xhline{2\arrayrulewidth}
\multicolumn{4}{|c|}{Baselines} \\ \hline
HyenaDNA \cite{hyenadna}                    & 36.95 & 35.38 & 72.87 \\ \hline
NT-2500M-multi \cite{dalla2023nucleotide}   & 70.33 & 71.58 & 72.97 \\ \hline
DNABERT-2 \cite{zhou2024dnabert2efficientfoundationmodel} & 69.37 & 68.04 & 74.17 \\ \hline
RandomMask \cite{randommask}                & 70.89 & 70.24 & 76.65 \\ \hline
LucaOne                                     & 60.82 & 66.93 & 75.19 \\ \hline
\end{tabular}
\end{table}

\clearpage

\paragraph*{S10 Table.}
\label{tape1}
\textbf{Secondary structure performance. In the 3-way columns, CASP12, CB513, and TS115 scores are reported; in the 8-way columns, the corresponding scores are reported. All values are measured in accuracy.}

\begin{table}[!h]
\centering
\small
\begin{tabular}{|l|ccc|ccc|}
\hline
\multirow{2}{*}{Model} & \multicolumn{3}{c|}{Secondary Structure (3-way)} & \multicolumn{3}{c|}{Secondary Structure (8-way)} \\ \cline{2-7} 
                       & CASP12 & CB513 & TS115             & CASP12 & CB513 & TS115             \\ \hline
\Xhline{2\arrayrulewidth}
\multicolumn{7}{|c|}{\textbf{OmniBioTE}} \\ \hline
OmniBioTE-small  & 0.695 & 0.733 & 0.762 & 0.568 & 0.598 & 0.640 \\ \hline
OmniBioTE-medium & 0.717 & 0.784 & 0.794 & 0.600 & 0.642 & 0.680 \\ \hline
OmniBioTE-large  & 0.722 & 0.786 & 0.801 & 0.591 & 0.646 & 0.674 \\ \hline
OmniBioTE-XL     & 0.708 & 0.798 & 0.805 & 0.582 & 0.656 & 0.681 \\ \hline
\Xhline{2\arrayrulewidth}
\multicolumn{7}{|c|}{\textbf{OmniBioTE (per-residue)}} \\ \hline
OmniBioTE-small (per-residue)  & 0.721 & 0.757 & 0.787 & 0.585 & 0.616 & 0.669 \\ \hline
OmniBioTE-medium (per-residue) & 0.746 & 0.813 & 0.820 & 0.619 & 0.678 & 0.707 \\ \hline
OmniBioTE-large (per-residue)  & 0.749 & 0.819 & 0.825 & 0.630 & 0.685 & 0.705 \\ \hline
OmniBioTE-XL (per-residue)     & 0.751 & 0.822 & 0.828 & 0.615 & 0.688 & 0.716 \\ \hline
\Xhline{2\arrayrulewidth}
\multicolumn{7}{|c|}{\textbf{ProtBioTE}} \\ \hline
ProtBioTE-small  & 0.707 & 0.769 & 0.782 & 0.568 & 0.626 & 0.667 \\ \hline
ProtBioTE-medium & 0.717 & 0.784 & 0.794 & 0.600 & 0.642 & 0.680 \\ \hline
ProtBioTE-large  & 0.767 & 0.822 & 0.828 & 0.591 & 0.646 & 0.674 \\ \hline
ProtBioTE-XL     & 0.764 & 0.827 & 0.831 & 0.642 & 0.691 & 0.717 \\ \hline
\Xhline{2\arrayrulewidth}
\multicolumn{7}{|c|}{\textbf{Baselines}} \\ \hline
ESM2-t6-8M      & 0.702 & 0.731 & 0.658 & 0.590 & 0.586 & 0.658 \\ \hline
ESM2-t12-35M    & 0.730 & 0.773 & 0.805 & 0.607 & 0.631 & 0.690 \\ \hline
ESM2-t30-150M   & 0.753 & 0.802 & 0.716 & 0.634 & 0.668 & 0.716 \\ \hline
ESM2-t33-650M   & 0.780 & 0.831 & 0.843 & 0.667 & 0.700 & 0.733 \\ \hline
ESM2-t36-3B     & 0.781 & 0.826 & 0.842 & 0.668 & 0.701 & 0.740 \\ \hline
LucaOne         & 0.700 & 0.720 & 0.755 & 0.578 & 0.569 & 0.630 \\ \hline
TAPE-Transformer& 0.710 & 0.730 & 0.770 & 0.590 & 0.590 & 0.640 \\ \hline
TAPE-ResNet     & 0.700 & 0.750 & 0.780 & 0.570 & 0.590 & 0.660 \\ \hline
TAPE-LSTM       & 0.720 & 0.750 & 0.780 & 0.580 & 0.590 & 0.640 \\ \hline
Supervised \cite{bepler2019learning} & 0.700 & 0.730 & 0.760 & 0.570 & 0.580 & 0.650 \\ \hline
UniRep \cite{alley2019unified}     & 0.720 & 0.730 & 0.770 & 0.590 & 0.570 & 0.630 \\ \hline
\end{tabular}
\end{table}

\clearpage

\paragraph*{S11 Table.}
\label{tape2}
\textbf{Remote homology (Fold, Superfamily, Family) classification performance measured in accuracy and regression performance (Fluorescence, Stability) measured in Spearman's correlation coefficient.}
\begin{table}[!h]
\centering
\small
\begin{tabular}{|l|c|c|c|c|c|}
\hline
Model & Fold & Superfamily & Family & Fluorescence & Stability \\ \hline
\Xhline{2\arrayrulewidth}
\multicolumn{6}{|c|}{\textbf{OmniBioTE}} \\ \hline
OmniBioTE-small  & 0.208 & 0.906 & 0.362 & 0.666 & 0.686 \\ \hline
OmniBioTE-medium & 0.219 & 0.965 & 0.454 & 0.655 & 0.722 \\ \hline
OmniBioTE-large  & 0.226 & 0.971 & 0.455 & 0.660 & 0.671 \\ \hline
OmniBioTE-XL     & 0.242 & 0.970 & 0.482 & 0.659 & 0.689 \\ \hline
\Xhline{2\arrayrulewidth}
\multicolumn{6}{|c|}{\textbf{OmniBioTE (per-residue)}} \\ \hline
OmniBioTE-small (per-residue)  & 0.201 & 0.914 & 0.342 & 0.659 & 0.700 \\ \hline
OmniBioTE-medium (per-residue) & 0.231 & 0.966 & 0.475 & 0.587 & 0.689 \\ \hline
OmniBioTE-large (per-residue)  & 0.240 & 0.972 & 0.512 & 0.662 & 0.711 \\ \hline
OmniBioTE-XL (per-residue)     & 0.223 & 0.973 & 0.470 & 0.539 & 0.699 \\ \hline
\Xhline{2\arrayrulewidth}
\multicolumn{6}{|c|}{\textbf{ProtBioTE}} \\ \hline
ProtBioTE-small  & 0.194 & 0.951 & 0.406 & 0.666 & 0.702 \\ \hline
ProtBioTE-medium & 0.219 & 0.965 & 0.454 & 0.655 & 0.722 \\ \hline
ProtBioTE-large  & 0.226 & 0.971 & 0.455 & 0.666 & 0.683 \\ \hline
ProtBioTE-XL     & 0.241 & 0.972 & 0.463 & 0.663 & 0.654 \\ \hline
\Xhline{2\arrayrulewidth}
\multicolumn{6}{|c|}{\textbf{Baselines}} \\ \hline
ESM2-t6-8M      & 0.240 & 0.911 & 0.439 & 0.663 & 0.660 \\ \hline
ESM2-t12-35M    & 0.288 & 0.961 & 0.574 & 0.673 & 0.723 \\ \hline
ESM2-t30-150M   & 0.272 & 0.978 & 0.601 & 0.672 & 0.761 \\ \hline
ESM2-t33-650M   & 0.231 & 0.965 & 0.530 & 0.665 & 0.720 \\ \hline
ESM2-t36-3B     & 0.249 & 0.970 & 0.542 & 0.654 & 0.774 \\ \hline
LucaOne         & 0.266 & 0.949 & 0.487 & 0.639 & 0.703 \\ \hline
TAPE-Transformer& 0.21  & 0.88  & 0.34  & 0.68  & 0.73  \\ \hline
TAPE-ResNet     & 0.26  & 0.92  & 0.43  & 0.67  & 0.69  \\ \hline
TAPE-LSTM       & 0.17  & 0.77  & 0.31  & 0.21  & 0.73  \\ \hline
Supervised \cite{bepler2019learning} & 0.17  & 0.79  & 0.20  & 0.33  & 0.64  \\ \hline
UniRep \cite{alley2019unified}     & 0.23  & 0.87  & 0.38  & 0.67  & 0.73  \\ \hline
\end{tabular}
\end{table}

\clearpage

\paragraph*{S12 Table.}
\label{tape3}
\textbf{Contact evaluation performance, reporting Contacts P@L for long- and medium-range contacts. All values are the computed precision of the predictions.}
\begin{table}[!h]
\centering
\small
\begin{tabular}{|l|c|c|}
\hline
Model & Contacts P@L (long) & Contacts P@L (medium) \\ \hline
\Xhline{2\arrayrulewidth}
\multicolumn{3}{|c|}{\textbf{OmniBioTE}} \\ \hline
OmniBioTE-small  & 0.286 & 0.339 \\ \hline
OmniBioTE-medium & 0.237 & 0.350 \\ \hline
OmniBioTE-large  & 0.280 & 0.371 \\ \hline
OmniBioTE-XL     & 0.300 & 0.334 \\ \hline
\Xhline{2\arrayrulewidth}
\multicolumn{3}{|c|}{\textbf{OmniBioTE (per-residue)}} \\ \hline
OmniBioTE-small (per-residue)  & 0.544 & 0.682 \\ \hline
OmniBioTE-medium (per-residue) & 0.467 & 0.614 \\ \hline
OmniBioTE-large (per-residue)  & 0.636 & 0.725 \\ \hline
OmniBioTE-XL (per-residue)     & 0.755 & 0.789 \\ \hline
\Xhline{2\arrayrulewidth}
\multicolumn{3}{|c|}{\textbf{ProtBioTE}} \\ \hline
ProtBioTE-small  & 0.307 & 0.373 \\ \hline
ProtBioTE-medium & 0.386 & 0.406 \\ \hline
ProtBioTE-large  & 0.302 & 0.347 \\ \hline
ProtBioTE-XL     & 0.318 & 0.394 \\ \hline
\Xhline{2\arrayrulewidth}
\multicolumn{3}{|c|}{\textbf{Baselines}} \\ \hline
ESM2-t6-8M      & 0.521 & 0.609 \\ \hline
ESM2-t12-35M    & 0.506 & 0.676 \\ \hline
ESM2-t30-150M   & 0.515 & 0.654 \\ \hline
ESM2-t33-650M   & 0.765 & 0.822 \\ \hline
ESM2-t36-3B     & 0.753 & 0.819 \\ \hline
LucaOne         & 0.365 & 0.556 \\ \hline
TAPE-Transformer& 0.17  & 0.19  \\ \hline
TAPE-ResNet     & 0.20  & 0.20  \\ \hline
TAPE-LSTM       & 0.10  & 0.18  \\ \hline
Supervised [11] & 0.18  & 0.22  \\ \hline
UniRep [12]     & 0.17  & 0.17  \\ \hline
\end{tabular}
\end{table}

\clearpage

\paragraph*{S13 Table.}
\label{protein_scores1}
\textbf{Performance on the structural prediction tasks in the ProteinGLUE dataset. Values represent the accuracy of the predictions.}

\begin{table}[!h]
\centering
\begin{tabular}{|l|r|r|r|r|}
\hline
Model & SS3 & SS8 & SS3 CB513 & SS8 CB513 \\ \hline
\Xhline{2\arrayrulewidth}
\multicolumn{5}{|c|}{OmniBioTE} \\ \hline
OmniBioTE-small & 77.1 & 64.9 & 77.6 & 63.0 \\ \hline
OmniBioTE-medium & 81.3 & 69.0 & 82.9 & 68.5 \\ \hline
OmniBioTE-large  & 82.0 & 69.8 & 83.4 & 69.7 \\ \hline
OmniBioTE-XL     & 82.7 & 70.7 & 87.0 & 72.0 \\ \hline
\Xhline{2\arrayrulewidth}
\multicolumn{5}{|c|}{OmniBioTE (per-residue)} \\ \hline
OmniBioTE-small (per-residue)  & 76.7 & 64.5 & 76.4 & 62.6 \\ \hline
OmniBioTE-medium (per-residue) & 81.8 & 69.4 & 82.9 & 69.9 \\ \hline
OmniBioTE-large (per-residue)  & 82.5 & 70.6 & 83.0 & 69.5 \\ \hline
OmniBioTE-XL (per-residue)     & 82.8 & 71.1 & 83.5 & 73.0 \\ \hline
\Xhline{2\arrayrulewidth}
\multicolumn{5}{|c|}{ProtBioTE} \\ \hline
ProtBioTE-small  & 79.8 & 67.3 & 81.3 & 67.0 \\ \hline
ProtBioTE-medium & 84.1 & 72.3 & 87.8 & 72.8 \\ \hline
ProtBioTE-large  & 84.9 & 73.0 & 86.5 & 73.8 \\ \hline
ProtBioTE-XL     & 85.4 & 74.3 & 88.8 & 75.0 \\ \hline
\Xhline{2\arrayrulewidth}
ESM2-t6-8M          & 76.0 & 63.8 & 73.4 & 58.7 \\ \hline
ESM2-t12-35M        & 79.9 & 67.8 & 77.3 & 63.0 \\ \hline
ESM2-t30-150M       & 83.0 & 71.7 & 81.0 & 67.6 \\ \hline
ESM2-t33-650M       & 85.3 & 83.0 & 83.0 & 70.4 \\ \hline
ESM2-t36-3B         & 85.6 & 75.1 & 82.8 & 70.5 \\ \hline
LucaOne             & 75.5 & 62.8 & 73.2 & 58.2 \\ \hline
\end{tabular}
\end{table}

\clearpage

\paragraph*{S14 Table.}
\label{protein_scores2}
\textbf{Performance on the remaining tasks in the ProteinGLUE dataset.}

\begin{table}[!h]
\centering
\resizebox{\textwidth}{!}{%
\begin{tabular}{|l|r|r|r|r|r|}
\hline
Model & 
\makecell{Protein-protein\\Interaction\\(AUCROC)} & 
\makecell{Hydrophobic\\patch rank\\(PCC)} & 
\makecell{Epitope\\detection\\(AUCROC)} & 
\makecell{Solvent\\accessibility\\(PCC)} & 
\makecell{Buried-residue\\prediction\\(Accuracy)} \\ 
\Xhline{2\arrayrulewidth}
\multicolumn{6}{|c|}{OmniBioTE} \\ \hline
OmniBioTE-small-1k & 0.580 & 0.264 & 0.502 & 0.615 & 86.6 \\ \hline
OmniBioTE-medium-1k & 0.618 & 0.239 & 0.616 & 0.657 & 88.0 \\ \hline
OmniBioTE-large-1k  & 0.591 & 0.237 & 0.559 & 0.642 & 87.9 \\ \hline
OmniBioTE-XL-1k     & 0.608 & 0.265 & 0.595 & 0.565 & 88.1 \\ \hline
\Xhline{2\arrayrulewidth}
\multicolumn{6}{|c|}{OmniBioTE (per-residue)} \\ \hline
OmniBioTE-small (per-residue)  & 0.532 & 0.145 & 0.628 & 0.598 & 80.0 \\ \hline
OmniBioTE-medium (per-residue) & 0.568 & 0.114 & 0.573 & 0.623 & 82.8 \\ \hline
OmniBioTE-large (per-residue)  & 0.641 & 0.133 & 0.554 & 0.659 & 83.3 \\ \hline
OmniBioTE-XL (per-residue)     & 0.610 & 0.075 & 0.590 & 0.617 & 83.3 \\ \hline
\Xhline{2\arrayrulewidth}
\multicolumn{6}{|c|}{ProtBioTE} \\ \hline
ProtBioTE-small-1k  & 0.585 & 0.272 & 0.536 & 0.624 & 87.2 \\ \hline
ProtBioTE-medium-1k & 0.619 & 0.304 & 0.515 & 0.654 & 88.6 \\ \hline
ProtBioTE-large-1k  & 0.599 & 0.074 & 0.610 & 0.554 & 88.8 \\ \hline
ProtBioTE-XL-1k     & 0.617 & 0.284 & 0.562 & 0.593 & 89.2 \\ \hline
\Xhline{2\arrayrulewidth}
ESM2-t6-8M          & 0.617 & 0.303 & 0.517 & 0.676 & 79.7 \\ \hline
ESM2-t12-35M        & 0.536 & 0.211 & 0.503 & 0.711 & 82.0 \\ \hline
ESM2-t30-150M       & 0.659 & 0.200 & 0.504 & 0.749 & 83.6 \\ \hline
ESM2-t33-650M       & 0.653 & 0.202 & 0.523 & 0.769 & 84.8 \\ \hline
ESM2-t36-3B         & 0.653 & 0.177 & 0.514 & 0.752 & 85.3 \\ \hline
LucaOne             & 0.597 & 0.275 & 0.502 & 0.663 & 79.4 \\ \hline
\end{tabular}
}
\end{table}

\clearpage  

\bibliography{ref}

@article{zhang2019unified,
  title={Unified efficient thermostat scheme for the canonical ensemble with holonomic or isokinetic constraints via molecular dynamics},
  author={Zhang, Zhijun and Liu, Xinzijian and Yan, Kangyu and Tuckerman, Mark E and Liu, Jian},
  journal={J Phys Chem A},
  volume={123},
  number={28},
  pages={6056--6079},
  year={2019},
  publisher={ACS Publications}
}

@article{nguyen2013improved,
  title={Improved generalized born solvent model parameters for protein simulations},
  author={Nguyen, Hai and Roe, Daniel R and Simmerling, Carlos},
  journal={J. Chem. Theor. Comput.},
  volume={9},
  number={4},
  pages={2020--2034},
  year={2013},
  publisher={ACS Publications}
}

@article{maier2015ff14sb,
  title={ff14SB: improving the accuracy of protein side chain and backbone parameters from ff99SB},
  author={Maier, James A and Martinez, Carmenza and Kasavajhala, Koushik and Wickstrom, Lauren and Hauser, Kevin E and Simmerling, Carlos},
  journal={J. Chem. Theor. Comput.},
  volume={11},
  number={8},
  pages={3696--3713},
  year={2015},
  publisher={ACS Publications}
}

@article{leimkuhler2013robust,
  title={Robust and efficient configurational molecular sampling via Langevin dynamics},
  author={Leimkuhler, Benedict and Matthews, Charles},
  journal={J Chem Phys},
  volume={138},
  number={17},
  year={2013},
  publisher={AIP Publishing}
}

@article{eastman2023openmm,
  title={OpenMM 8: molecular dynamics simulation with machine learning potentials},
  author={Eastman, Peter and Galvelis, Raimondas and Pel{\'a}ez, Ra{\'u}l P and Abreu, Charlles RA and Farr, Stephen E and Gallicchio, Emilio and Gorenko, Anton and Henry, Michael M and Hu, Frank and Huang, Jing and others},
  journal={J. Phys. Chem. B},
  volume={128},
  number={1},
  pages={109--116},
  year={2023},
  publisher={ACS Publications}
}

@article{roux2024editorial,
  title={Editorial Guidelines for Computational Studies of Ligand Binding Using MM/PBSA and MM/GBSA Approximations Wisely},
  author={Roux, Beno{\^\i}t and Chipot, Christophe},
  journal={J. Phys. Chem. B},
  volume={128},
  number={49},
  pages={12027--12029},
  year={2024},
  publisher={ACS Publications}
}

@article{wang2019end,
  title={End-point binding free energy calculation with MM/PBSA and MM/GBSA: strategies and applications in drug design},
  author={Wang, Ercheng and Sun, Huiyong and Wang, Junmei and Wang, Zhe and Liu, Hui and Zhang, John ZH and Hou, Tingjun},
  journal={Chem. Rev.},
  volume={119},
  number={16},
  pages={9478--9508},
  year={2019},
  publisher={ACS Publications}
}

@article {evo2,
	author = {Brixi, Garyk and Durrant, Matthew G and Ku, Jerome and Poli, Michael and Brockman, Greg and Chang, Daniel and Gonzalez, Gabriel A and King, Samuel H and Li, David B and Merchant, Aditi T and Naghipourfar, Mohsen and Nguyen, Eric and Ricci-Tam, Chiara and Romero, David W and Sun, Gwanggyu and Taghibakshi, Ali and Vorontsov, Anton and Yang, Brandon and Deng, Myra and Gorton, Liv and Nguyen, Nam and Wang, Nicholas K and Adams, Etowah and Baccus, Stephen A and Dillmann, Steven and Ermon, Stefano and Guo, Daniel and Ilango, Rajesh and Janik, Ken and Lu, Amy X and Mehta, Reshma and Mofrad, Mohammad R.K. and Ng, Madelena Y and Pannu, Jaspreet and Re, Christopher and Schmok, Jonathan C and St. John, John and Sullivan, Jeremy and Zhu, Kevin and Zynda, Greg and Balsam, Daniel and Collison, Patrick and Costa, Anthony B. and Hernandez-Boussard, Tina and Ho, Eric and Liu, Ming-Yu and McGrath, Tom and Powell, Kimberly and Burke, Dave P. and Goodarzi, Hani and Hsu, Patrick D and Hie, Brian},
	title = {Genome modeling and design across all domains of life with Evo 2},
	elocation-id = {2025.02.18.638918},
	year = {2025},
	doi = {10.1101/2025.02.18.638918},
	publisher = {Cold Spring Harbor Laboratory},
	URL = {https://www.biorxiv.org/content/early/2025/02/21/2025.02.18.638918},
	eprint = {https://www.biorxiv.org/content/early/2025/02/21/2025.02.18.638918.full.pdf},
	journal = {bioRxiv}
}

@article{10.1093/nar/gky989,
    author = {Sayers, Eric W and Cavanaugh, Mark and Clark, Karen and Ostell, James and Pruitt, Kim D and Karsch-Mizrachi, Ilene},
    title = "{GenBank}",
    journal = {Nucleic Acids Research},
    volume = {47},
    number = {D1},
    pages = {D94-D99},
    year = {2018},
    month = {10},
    issn = {0305-1048},
    doi = {10.1093/nar/gky989},
    url = {https://doi.org/10.1093/nar/gky989},
    eprint = {https://academic.oup.com/nar/article-pdf/47/D1/D94/27436415/gky989.pdf},
}

@misc{kudo2018sentencepiecesimplelanguageindependent,
      title={SentencePiece: A simple and language independent subword tokenizer and detokenizer for Neural Text Processing}, 
      author={Taku Kudo and John Richardson},
      year={2018},
      eprint={1808.06226},
      archivePrefix={arXiv},
      primaryClass={cs.CL},
      url={https://arxiv.org/abs/1808.06226}, 
}

@misc{zhou2024dnabert2efficientfoundationmodel,
      title={DNABERT-2: Efficient Foundation Model and Benchmark For Multi-Species Genome}, 
      author={Zhihan Zhou and Yanrong Ji and Weijian Li and Pratik Dutta and Ramana Davuluri and Han Liu},
      year={2024},
      eprint={2306.15006},
      archivePrefix={arXiv},
      primaryClass={q-bio.GN},
      url={https://arxiv.org/abs/2306.15006}, 
}

@article{radford2019language,
  title={Language Models are Unsupervised Multitask Learners},
  author={Radford, Alec and Wu, Jeff and Child, Rewon and Luan, David and Amodei, Dario and Sutskever, Ilya},
  year={2019}
}

@misc{yang2022tensorprogramsvtuning,
      title={Tensor Programs V: Tuning Large Neural Networks via Zero-Shot Hyperparameter Transfer}, 
      author={Greg Yang and Edward J. Hu and Igor Babuschkin and Szymon Sidor and Xiaodong Liu and David Farhi and Nick Ryder and Jakub Pachocki and Weizhu Chen and Jianfeng Gao},
      year={2022},
      eprint={2203.03466},
      archivePrefix={arXiv},
      primaryClass={cs.LG},
      url={https://arxiv.org/abs/2203.03466}, 
}

@misc{touvron2023llama2openfoundation,
      title={Llama 2: Open Foundation and Fine-Tuned Chat Models}, 
      author={Hugo Touvron and Louis Martin and Kevin Stone and Peter Albert and Amjad Almahairi and Yasmine Babaei and Nikolay Bashlykov and Soumya Batra and Prajjwal Bhargava and Shruti Bhosale and Dan Bikel and Lukas Blecher and Cristian Canton Ferrer and Moya Chen and Guillem Cucurull and David Esiobu and Jude Fernandes and Jeremy Fu and Wenyin Fu and Brian Fuller and Cynthia Gao and Vedanuj Goswami and Naman Goyal and Anthony Hartshorn and Saghar Hosseini and Rui Hou and Hakan Inan and Marcin Kardas and Viktor Kerkez and Madian Khabsa and Isabel Kloumann and Artem Korenev and Punit Singh Koura and Marie-Anne Lachaux and Thibaut Lavril and Jenya Lee and Diana Liskovich and Yinghai Lu and Yuning Mao and Xavier Martinet and Todor Mihaylov and Pushkar Mishra and Igor Molybog and Yixin Nie and Andrew Poulton and Jeremy Reizenstein and Rashi Rungta and Kalyan Saladi and Alan Schelten and Ruan Silva and Eric Michael Smith and Ranjan Subramanian and Xiaoqing Ellen Tan and Binh Tang and Ross Taylor and Adina Williams and Jian Xiang Kuan and Puxin Xu and Zheng Yan and Iliyan Zarov and Yuchen Zhang and Angela Fan and Melanie Kambadur and Sharan Narang and Aurelien Rodriguez and Robert Stojnic and Sergey Edunov and Thomas Scialom},
      year={2023},
      eprint={2307.09288},
      archivePrefix={arXiv},
      primaryClass={cs.CL},
      url={https://arxiv.org/abs/2307.09288}, 
}

@misc{Karpathy2022,
  author = {Andrej Karpathy},
  title = {\text{NanoGPT}},
  year = {2022},
  publisher = {GitHub},
  journal = {GitHub repository},
  howpublished = {\url{https://github.com/karpathy/nanoGPT}},
  commit = {325be85d9be8c81b436728a420e85796c57dba7e}
}

@misc{devlin2019bertpretrainingdeepbidirectional,
      title={BERT: Pre-training of Deep Bidirectional Transformers for Language Understanding}, 
      author={Jacob Devlin and Ming-Wei Chang and Kenton Lee and Kristina Toutanova},
      year={2019},
      eprint={1810.04805},
      archivePrefix={arXiv},
      primaryClass={cs.CL},
      url={https://arxiv.org/abs/1810.04805}, 
}

@misc{loshchilov2019decoupledweightdecayregularization,
      title={Decoupled Weight Decay Regularization}, 
      author={Ilya Loshchilov and Frank Hutter},
      year={2019},
      eprint={1711.05101},
      archivePrefix={arXiv},
      primaryClass={cs.LG},
      url={https://arxiv.org/abs/1711.05101}, 
}

@misc{smith2018superconvergencefasttrainingneural,
      title={Super-Convergence: Very Fast Training of Neural Networks Using Large Learning Rates}, 
      author={Leslie N. Smith and Nicholay Topin},
      year={2018},
      eprint={1708.07120},
      archivePrefix={arXiv},
      primaryClass={cs.LG},
      url={https://arxiv.org/abs/1708.07120}, 
}

@article{tape,
  title={Evaluating protein transfer learning with TAPE},
  author={Rao, Roshan and Bhattacharya, Nicholas and Thomas, Neil and Duan, Yan and Chen, Peter and Canny, John and Abbeel, Pieter and Song, Yun},
  journal={Advances in neural information processing systems},
  volume={32},
  year={2019}
}

@misc{he2015deepresiduallearningimage,
      title={Deep Residual Learning for Image Recognition}, 
      author={Kaiming He and Xiangyu Zhang and Shaoqing Ren and Jian Sun},
      year={2015},
      eprint={1512.03385},
      archivePrefix={arXiv},
      primaryClass={cs.CV},
      url={https://arxiv.org/abs/1512.03385}, 
}

@article {Capel2021.12.13.472460,
	author = {Capel, Henriette and Weiler, Robin and Dijkstra, Maurits and Vleugels, Reinier and Bloem, Peter and Feenstra, K. Anton},
	title = {ProteinGLUE: A multi-task benchmark suite for self-supervised protein modeling},
	elocation-id = {2021.12.13.472460},
	year = {2021},
	doi = {10.1101/2021.12.13.472460},
	publisher = {Cold Spring Harbor Laboratory},
	URL = {https://www.biorxiv.org/content/early/2021/12/14/2021.12.13.472460},
	eprint = {https://www.biorxiv.org/content/early/2021/12/14/2021.12.13.472460.full.pdf},
	journal = {bioRxiv}
}

@article{10.1093/nar/gkab848,
    author = {Harini, Kannan and Srivastava, Ambuj and Kulandaisamy, Arulsamy and Gromiha, M Michael},
    title = "{ProNAB: database for binding affinities of protein–nucleic acid complexes and their mutants}",
    journal = {Nucleic Acids Research},
    volume = {50},
    number = {D1},
    pages = {D1528-D1534},
    year = {2021},
    month = {10},
    issn = {0305-1048},
    doi = {10.1093/nar/gkab848},
    url = {https://doi.org/10.1093/nar/gkab848},
    eprint = {https://academic.oup.com/nar/article-pdf/50/D1/D1528/42058446/gkab848.pdf},
}

@misc{sennrich2016neuralmachinetranslationrare,
      title={Neural Machine Translation of Rare Words with Subword Units}, 
      author={Rico Sennrich and Barry Haddow and Alexandra Birch},
      year={2016},
      eprint={1508.07909},
      archivePrefix={arXiv},
      primaryClass={cs.CL},
      url={https://arxiv.org/abs/1508.07909}, 
}

@article{10.1093/bioinformatics/btm098,
    author = {Suzek, Baris E. and Huang, Hongzhan and McGarvey, Peter and Mazumder, Raja and Wu, Cathy H.},
    title = "{UniRef: comprehensive and non-redundant UniProt reference clusters}",
    journal = {Bioinformatics},
    volume = {23},
    number = {10},
    pages = {1282-1288},
    year = {2007},
    month = {03},
    issn = {1367-4803},
    doi = {10.1093/bioinformatics/btm098},
    url = {https://doi.org/10.1093/bioinformatics/btm098},
    eprint = {https://academic.oup.com/bioinformatics/article-pdf/23/10/1282/49812789/bioinformatics\_23\_10\_1282.pdf},
}

@article{ma2015protein,
  title={Protein contact prediction by integrating joint evolutionary coupling analysis and supervised learning},
  author={Ma, Jianzhu and Wang, Sheng and Wang, Zhiyong and Xu, Jinbo},
  journal={Bioinformatics},
  volume={31},
  number={21},
  pages={3506--3513},
  year={2015},
  publisher={Oxford University Press}
}

@article{10.1093/nar/28.1.235,
    author = {Berman, Helen M. and Westbrook, John and Feng, Zukang and Gilliland, Gary and Bhat, T. N. and Weissig, Helge and Shindyalov, Ilya N. and Bourne, Philip E.},
    title = "{The Protein Data Bank}",
    journal = {Nucleic Acids Research},
    volume = {28},
    number = {1},
    pages = {235-242},
    year = {2000},
    month = {01},
    issn = {0305-1048},
    doi = {10.1093/nar/28.1.235},
    url = {https://doi.org/10.1093/nar/28.1.235},
    eprint = {https://academic.oup.com/nar/article-pdf/28/1/235/9895144/280235.pdf},
}

@article{dalla2023nucleotide,
  title={The Nucleotide Transformer: Building and Evaluating Robust Foundation Models for Human Genomics},
  author={Dalla-Torre, Hugo and Gonzalez, Liam and Mendoza Revilla, Javier and Lopez Carranza, Nicolas and Henryk Grywaczewski, Adam and Oteri, Francesco and Dallago, Christian and Trop, Evan and Sirelkhatim, Hassan and Richard, Guillaume and others},
  journal={bioRxiv},
  pages={2023--01},
  year={2023},
  publisher={Cold Spring Harbor Laboratory}
}

@inproceedings{transformer,
author = {Vaswani, Ashish and Shazeer, Noam and Parmar, Niki and Uszkoreit, Jakob and Jones, Llion and Gomez, Aidan N. and Kaiser, \L{}ukasz and Polosukhin, Illia},
title = {Attention is all you need},
year = {2017},
isbn = {9781510860964},
publisher = {Curran Associates Inc.},
address = {Red Hook, NY, USA},
booktitle = {Proceedings of the 31st International Conference on Neural Information Processing Systems},
pages = {6000–6010},
numpages = {11},
location = {Long Beach, California, USA},
series = {NIPS'17}
}

@article{pegaptanib,
author = {Evangelos S. Gragoudas  and Anthony P. Adamis  and Emmett T. Cunningham  and Matthew Feinsod  and David R. Guyer },
title = {Pegaptanib for Neovascular Age-Related Macular Degeneration},
journal = {New England Journal of Medicine},
volume = {351},
number = {27},
pages = {2805-2816},
year = {2004},
doi = {10.1056/NEJMoa042760},

URL = {https://www.nejm.org/doi/full/10.1056/NEJMoa042760},
eprint = {https://www.nejm.org/doi/pdf/10.1056/NEJMoa042760}
}

@article{Carvalho2019,
  author = {Josué Carvalho and Artur Paiva and Maria Paula Cabral Campello and António Paulo and Jean-Louis Mergny and Gilmar F. Salgado and João A. Queiroz and Carla Cruz},
  title = {Aptamer-based Targeted Delivery of a G-quadruplex Ligand in Cervical Cancer Cells},
  journal = {Scientific Reports},
  year = {2019},
  volume = {9},
  number = {1},
  pages = {7945},
  doi = {10.1038/s41598-019-44388-9},
  url = {https://doi.org/10.1038/s41598-019-44388-9}
}

@article{Tanaka2009,
  author = {Kenichi A. Tanaka and Fania Szlam and Christopher P. Rusconi and Jerrold H. Levy},
  title = {In-vitro evaluation of anti-factor IXa aptamer on thrombin generation, clotting time, and viscoelastometry},
  journal = {Thrombosis and Haemostasis},
  year = {2009},
  volume = {101},
  number = {5},
  pages = {827-833},
  month = {May},
  issn = {0340-6245},
  pmid = {19404534},
  doi = {10.1055/s-0037-1618602},
  url = {https://www.thieme-connect.com/products/ejournals/abstract/10.1055/s-0037-1618602},
}

@article{Chan2008,
  author = {M. Y. Chan and C. P. Rusconi and J. H. Alexander and R. M. Tonkens and R. A. Harrington and R. C. Becker},
  title = {A randomized, repeat-dose, pharmacodynamic and safety study of an antidote-controlled factor IXa inhibitor},
  journal = {Journal of Thrombosis and Haemostasis},
  year = {2008},
  volume = {6},
  number = {5},
  pages = {789-796},
  month = {May},
  doi = {10.1111/j.1538-7836.2008.02932.x},
  pmid = {18284597},
  issn = {1538-7836},
  url = {https://doi.org/10.1111/j.1538-7836.2008.02932.x},
}

@article{Riccardi2020,
  author = {Claudia Riccardi and Albert Meyer and Jean-Jacques Vasseur and Domenico Cavasso and Irene Russo Krauss and Luigi Paduano and François Morvan and Daniela Montesarchio},
  title = {Design, Synthesis and Characterization of Cyclic NU172 Analogues: A Biophysical and Biological Insight},
  journal = {International Journal of Molecular Sciences},
  year = {2020},
  volume = {21},
  number = {11},
  pages = {3860},
  month = {May},
  doi = {10.3390/ijms21113860},
  pmid = {32485818},
  issn = {1422-0067},
  url = {https://doi.org/10.3390/ijms21113860},
}

@article{JilmaStohlawetz2012,
  author = {Petra Jilma-Stohlawetz and Paul Knöbl and James C. Gilbert and Bernd Jilma},
  title = {The anti-von Willebrand factor aptamer ARC1779 increases von Willebrand factor levels and platelet counts in patients with type 2B von Willebrand disease},
  journal = {Thrombosis and Haemostasis},
  year = {2012},
  volume = {108},
  number = {2},
  pages = {284-290},
  month = {August},
  doi = {10.1160/TH11-12-0889},
  pmid = {22740102},
  issn = {0340-6245},
  url = {https://doi.org/10.1160/TH11-12-0889},
}

@article{Menne2017,
  author    = {Jan Menne and Dirk Eulberg and Diana Beyer and Matthias Baumann and Frantisek Saudek and Zsuzsanna Valkusz and Andrzej Więcek and Hermann Haller},
  title     = {C-C motif-ligand 2 inhibition with emapticap pegol (NOX-E36) in type 2 diabetic patients with albuminuria},
  journal   = {Nephrology, Dialysis, Transplantation},
  year      = {2017},
  volume    = {32},
  number    = {2},
  pages     = {307--315},
  doi       = {10.1093/ndt/gfv459},
  pmid      = {28186566},
  pmcid     = {PMC5410979},
  publisher = {Oxford University Press},
  issn      = {1460-2385},
}

@article{Giordano2024,
  author    = {Frank A. Giordano and Julian P. Layer and Sonia Leonardelli and Lea L. Friker and Roberta Turiello and Dillon Corvino and Thomas Zeyen and Christina Schaub and Wolf Müller and Elena Sperk and Leonard Christopher Schmeel and Katharina Sahm and Christoph Oster and Sied Kebir and Peter Hambsch and Torsten Pietsch and Sotirios Bisdas and Michael Platten and Martin Glas and Clemens Seidel and Ulrich Herrlinger and Michael Hölzel},
  title     = {L-RNA aptamer-based CXCL12 inhibition combined with radiotherapy in newly-diagnosed glioblastoma: dose escalation of the phase I/II GLORIA trial},
  journal   = {Nature Communications},
  year      = {2024},
  volume    = {15},
  number    = {1},
  pages     = {4210},
  doi       = {10.1038/s41467-024-48416-9},
  url       = {https://doi.org/10.1038/s41467-024-48416-9},
  issn      = {2041-1723},
}

@article{Schwoebel2013,
  author    = {Frank Schwoebel and Lucas T. van Eijk and Dirk Zboralski and Simone Sell and Klaus Buchner and Christian Maasch and Werner G. Purschke and Martin Humphrey and Stefan Zöllner and Dirk Eulberg and Frank Morich and Peter Pickkers and Sven Klussmann},
  title     = {The effects of the anti-hepcidin Spiegelmer NOX-H94 on inflammation-induced anemia in cynomolgus monkeys},
  journal   = {Blood},
  year      = {2013},
  volume    = {121},
  number    = {12},
  pages     = {2311--2315},
  doi       = {10.1182/blood-2012-09-456756},
  pmid      = {23349391},
  pmcid     = {PMC3606066},
  issn      = {1528-0020},
  eprint    = {2013 Jan 24},
  publisher = {American Society of Hematology},
}

@article{loh2014efficient,
  title={Efficient coding hypothesis and an introduction to information theory},
  author={Loh, Lay Kuan and Bartulovic, Mihovil},
  journal={Retrieved from users. ece. cmu. edu/\~{} pgrover/teaching/files/InfoTheoryEfficientCo dingHypothesis. pdf. Homayoun Shahri},
  year={2014}
}

@article{alphafold1,
  title={Improved protein structure prediction using potentials from deep learning},
  author={Senior, Andrew W and Evans, Richard and Jumper, John and Kirkpatrick, James and Sifre, Laurent and Green, Tim and Qin, Chongli and {\v{Z}}{\'\i}dek, Augustin and Nelson, Alexander WR and Bridgland, Alex and others},
  journal={Nature},
  volume={577},
  number={7792},
  pages={706--710},
  year={2020},
  publisher={Nature Publishing Group UK London}
}

@article{alphafold2,
  author = {John Jumper and Richard Evans and Alexander Pritzel and Tim Green and Michael Figurnov and Olaf Ronneberger and Kathryn Tunyasuvunakool and Russ Bates and Augustin Žídek and Anna Potapenko and Alex Bridgland and Clemens Meyer and Simon A. A. Kohl and Andrew J. Ballard and Andrew Cowie and Bernardino Romera-Paredes and Stanislav Nikolov and Rishub Jain and Jonas Adler and Trevor Back and Stig Petersen and David Reiman and Ellen Clancy and Michal Zielinski and Martin Steinegger and Michalina Pacholska and Tamas Berghammer and Sebastian Bodenstein and David Silver and Oriol Vinyals and Andrew W. Senior and Koray Kavukcuoglu and Pushmeet Kohli and Demis Hassabis},
  title = {Highly accurate protein structure prediction with AlphaFold},
  journal = {Nature},
  volume = {596},
  number = {7873},
  pages = {583--589},
  year = {2021},
  month = {aug},
  doi = {10.1038/s41586-021-03819-2},
  url = {https://doi.org/10.1038/s41586-021-03819-2},
  issn = {1476-4687},
}

@article{alphafold3,
  author = {Josh Abramson and Jonas Adler and Jack Dunger and Richard Evans and Tim Green and Alexander Pritzel and Olaf Ronneberger and Lindsay Willmore and Andrew J. Ballard and Joshua Bambrick and Sebastian W. Bodenstein and David A. Evans and Chia-Chun Hung and Michael O’Neill and David Reiman and Kathryn Tunyasuvunakool and Zachary Wu and Akvilė Žemgulytė and Eirini Arvaniti and Charles Beattie and Ottavia Bertolli and Alex Bridgland and Alexey Cherepanov and Miles Congreve and Alexander I. Cowen-Rivers and Andrew Cowie and Michael Figurnov and Fabian B. Fuchs and Hannah Gladman and Rishub Jain and Yousuf A. Khan and Caroline M. R. Low and Kuba Perlin and Anna Potapenko and Pascal Savy and Sukhdeep Singh and Adrian Stecula and Ashok Thillaisundaram and Catherine Tong and Sergei Yakneen and Ellen D. Zhong and Michal Zielinski and Augustin Žídek and Victor Bapst and Pushmeet Kohli and Max Jaderberg and Demis Hassabis and John M. Jumper},
  title = {Accurate structure prediction of biomolecular interactions with AlphaFold 3},
  journal = {Nature},
  volume = {630},
  number = {8016},
  pages = {493--500},
  year = {2024},
  month = {jun},
  doi = {10.1038/s41586-024-07487-w},
  url = {https://doi.org/10.1038/s41586-024-07487-w},
  issn = {1476-4687},
}

@article{rosettafold,
author = {Minkyung Baek  and Frank DiMaio  and Ivan Anishchenko  and Justas Dauparas  and Sergey Ovchinnikov  and Gyu Rie Lee  and Jue Wang  and Qian Cong  and Lisa N. Kinch  and R. Dustin Schaeffer  and Claudia Millán  and Hahnbeom Park  and Carson Adams  and Caleb R. Glassman  and Andy DeGiovanni  and Jose H. Pereira  and Andria V. Rodrigues  and Alberdina A. van Dijk  and Ana C. Ebrecht  and Diederik J. Opperman  and Theo Sagmeister  and Christoph Buhlheller  and Tea Pavkov-Keller  and Manoj K. Rathinaswamy  and Udit Dalwadi  and Calvin K. Yip  and John E. Burke  and K. Christopher Garcia  and Nick V. Grishin  and Paul D. Adams  and Randy J. Read  and David Baker },
title = {Accurate prediction of protein structures and interactions using a three-track neural network},
journal = {Science},
volume = {373},
number = {6557},
pages = {871-876},
year = {2021},
doi = {10.1126/science.abj8754},
URL = {https://www.science.org/doi/abs/10.1126/science.abj8754},
eprint = {https://www.science.org/doi/pdf/10.1126/science.abj8754}
}

@article{rosettafoldna,
  author = {Minkyung Baek and Ryan McHugh and Ivan Anishchenko and Hanlun Jiang and David Baker and Frank DiMaio},
  title = {Accurate prediction of protein–nucleic acid complexes using RoseTTAFoldNA},
  journal = {Nature Methods},
  volume = {21},
  number = {1},
  pages = {117--121},
  year = {2024},
  month = {jan},
  doi = {10.1038/s41592-023-02086-5},
  url = {https://doi.org/10.1038/s41592-023-02086-5},
  issn = {1548-7105},
}

@article{openfold,
  author = {Gustaf Ahdritz and Nazim Bouatta and Christina Floristean and Sachin Kadyan and Qinghui Xia and William Gerecke and Timothy J. O’Donnell and Daniel Berenberg and Ian Fisk and Niccolò Zanichelli and Bo Zhang and Arkadiusz Nowaczynski and Bei Wang and Marta M. Stepniewska-Dziubinska and Shang Zhang and Adegoke Ojewole and Murat Efe Guney and Stella Biderman and Andrew M. Watkins and Stephen Ra and Pablo Ribalta Lorenzo and Lucas Nivon and Brian Weitzner and Yih-En Andrew Ban and Shiyang Chen and Minjia Zhang and Conglong Li and Shuaiwen Leon Song and Yuxiong He and Peter K. Sorger and Emad Mostaque and Zhao Zhang and Richard Bonneau and Mohammed AlQuraishi},
  title = {OpenFold: retraining AlphaFold2 yields new insights into its learning mechanisms and capacity for generalization},
  journal = {Nature Methods},
  year = {2024},
  month = {may},
  doi = {10.1038/s41592-024-02272-z},
  url = {https://doi.org/10.1038/s41592-024-02272-z},
  issn = {1548-7105},
}

@article {omegafold,
	author = {Wu, Ruidong and Ding, Fan and Wang, Rui and Shen, Rui and Zhang, Xiwen and Luo, Shitong and Su, Chenpeng and Wu, Zuofan and Xie, Qi and Berger, Bonnie and Ma, Jianzhu and Peng, Jian},
	title = {High-resolution de novo structure prediction from primary sequence},
	elocation-id = {2022.07.21.500999},
	year = {2022},
	doi = {10.1101/2022.07.21.500999},
	publisher = {Cold Spring Harbor Laboratory},
	URL = {https://www.biorxiv.org/content/early/2022/07/22/2022.07.21.500999},
	eprint = {https://www.biorxiv.org/content/early/2022/07/22/2022.07.21.500999.full.pdf},
	journal = {bioRxiv}
}

@article{esmfold,
author = {Zeming Lin  and Halil Akin  and Roshan Rao  and Brian Hie  and Zhongkai Zhu  and Wenting Lu  and Nikita Smetanin  and Robert Verkuil  and Ori Kabeli  and Yaniv Shmueli  and Allan dos Santos Costa  and Maryam Fazel-Zarandi  and Tom Sercu  and Salvatore Candido  and Alexander Rives },
title = {Evolutionary-scale prediction of atomic-level protein structure with a language model},
journal = {Science},
volume = {379},
number = {6637},
pages = {1123-1130},
year = {2023},
doi = {10.1126/science.ade2574},
URL = {https://www.science.org/doi/abs/10.1126/science.ade2574},
eprint = {https://www.science.org/doi/pdf/10.1126/science.ade2574}
}

@article{enformer,
  author = {Žiga Avsec and Vikram Agarwal and Daniel Visentin and Joseph R. Ledsam and Agnieszka Grabska-Barwinska and Kyle R. Taylor and Yannis Assael and John Jumper and Pushmeet Kohli and David R. Kelley},
  title = {Effective gene expression prediction from sequence by integrating long-range interactions},
  journal = {Nature Methods},
  volume = {18},
  number = {10},
  pages = {1196--1203},
  year = {2021},
  month = {oct},
  doi = {10.1038/s41592-021-01252-x},
  url = {https://doi.org/10.1038/s41592-021-01252-x},
  issn = {1548-7105},
}

@article{genalm,
  title={Gena-lm: A family of open-source foundational models for long dna sequences},
  author={Fishman, Veniamin and Kuratov, Yuri and Petrov, Maxim and Shmelev, Aleksei and Shepelin, Denis and Chekanov, Nikolay and Kardymon, Olga and Burtsev, Mikhail},
  journal={bioRxiv},
  pages={2023--06},
  year={2023},
  publisher={Cold Spring Harbor Laboratory}
}

@article{genslm,
  title={GenSLMs: Genome-scale language models reveal SARS-CoV-2 evolutionary dynamics},
  author={Zvyagin, Maxim and Brace, Alexander and Hippe, Kyle and Deng, Yuntian and Zhang, Bin and Bohorquez, Cindy Orozco and Clyde, Austin and Kale, Bharat and Perez-Rivera, Danilo and Ma, Heng and others},
  journal={The International Journal of High Performance Computing Applications},
  volume={37},
  number={6},
  pages={683--705},
  year={2023},
  publisher={SAGE Publications Sage UK: London, England}
}

@article{hwang2024,
  title={Genomic language model predicts protein co-regulation and function},
  author={Hwang, Yunha and Cornman, Andre L and Kellogg, Elizabeth H and Ovchinnikov, Sergey and Girguis, Peter R},
  journal={Nature communications},
  volume={15},
  number={1},
  pages={2880},
  year={2024},
  publisher={Nature Publishing Group UK London}
}

@article{hyenadna,
  title={Hyenadna: Long-range genomic sequence modeling at single nucleotide resolution},
  author={Nguyen, Eric and Poli, Michael and Faizi, Marjan and Thomas, Armin and Wornow, Michael and Birch-Sykes, Callum and Massaroli, Stefano and Patel, Aman and Rabideau, Clayton and Bengio, Yoshua and others},
  journal={Advances in neural information processing systems},
  volume={36},
  year={2024}
}

@article{puffin,
  title={Sequence basis of transcription initiation in the human genome},
  author={Dudnyk, Kseniia and Cai, Donghong and Shi, Chenlai and Xu, Jian and Zhou, Jian},
  journal={Science},
  volume={384},
  number={6694},
  pages={eadj0116},
  year={2024},
  publisher={American Association for the Advancement of Science}
}

@article{bigrna,
  title={An RNA foundation model enables discovery of disease mechanisms and candidate therapeutics},
  author={Celaj, Albi and Gao, Alice Jiexin and Lau, Tammy TY and Holgersen, Erle M and Lo, Alston and Lodaya, Varun and Cole, Christopher B and Denroche, Robert E and Spickett, Carl and Wagih, Omar and others},
  journal={bioRxiv},
  pages={2023--09},
  year={2023},
  publisher={Cold Spring Harbor Laboratory}
}

@article{codonbert,
  title={Codonbert: Large language models for mrna design and optimization},
  author={Li, Sizhen and Moayedpour, Saeed and Li, Ruijiang and Bailey, Michael and Riahi, Saleh and Kogler-Anele, Lorenzo and Miladi, Milad and Miner, Jacob and Zheng, Dinghai and Wang, Jun and others},
  journal={bioRxiv},
  pages={2023--09},
  year={2023},
  publisher={Cold Spring Harbor Laboratory}
}

@article{theodoris2023transfer,
  title={Transfer learning enables predictions in network biology},
  author={Theodoris, Christina V and Xiao, Ling and Chopra, Anant and Chaffin, Mark D and Al Sayed, Zeina R and Hill, Matthew C and Mantineo, Helene and Brydon, Elizabeth M and Zeng, Zexian and Liu, X Shirley and others},
  journal={Nature},
  volume={618},
  number={7965},
  pages={616--624},
  year={2023},
  publisher={Nature Publishing Group UK London}
}

@article{fu2023get,
  title={GET: a foundation model of transcription across human cell types},
  author={Fu, Xi and Mo, Shentong and Buendia, Alejandro and Laurent, Anouchka and Shao, Anqi and Alvarez-Torres, Maria del Mar and Yu, Tianji and Tan, Jimin and Su, Jiayu and Sagatelian, Romella and others},
  journal={bioRxiv},
  pages={2023--09},
  year={2023},
  publisher={Cold Spring Harbor Laboratory}
}

@article{yang2022scbert,
  title={scBERT as a large-scale pretrained deep language model for cell type annotation of single-cell RNA-seq data},
  author={Yang, Fan and Wang, Wenchuan and Wang, Fang and Fang, Yuan and Tang, Duyu and Huang, Junzhou and Lu, Hui and Yao, Jianhua},
  journal={Nature Machine Intelligence},
  volume={4},
  number={10},
  pages={852--866},
  year={2022},
  publisher={Nature Publishing Group UK London}
}

@article{hao2024large,
  title={Large-scale foundation model on single-cell transcriptomics},
  author={Hao, Minsheng and Gong, Jing and Zeng, Xin and Liu, Chiming and Guo, Yucheng and Cheng, Xingyi and Wang, Taifeng and Ma, Jianzhu and Zhang, Xuegong and Song, Le},
  journal={Nature Methods},
  pages={1--11},
  year={2024},
  publisher={Nature Publishing Group US New York}
}

@article{cui2024scgpt,
  title={scGPT: toward building a foundation model for single-cell multi-omics using generative AI},
  author={Cui, Haotian and Wang, Chloe and Maan, Hassaan and Pang, Kuan and Luo, Fengning and Duan, Nan and Wang, Bo},
  journal={Nature Methods},
  pages={1--11},
  year={2024},
  publisher={Nature Publishing Group US New York}
}

@article{liu2023pre,
  title={A pre-trained large generative model for translating single-cell transcriptome to proteome},
  author={Liu, Linjing and Li, Wei and Wong, Ka-Chun and Yang, Fan and Yao, Jianhua},
  journal={bioRxiv},
  pages={2023--07},
  year={2023},
  publisher={Cold Spring Harbor Laboratory}
}

@article{shen2023generative,
  title={Generative pretraining from large-scale transcriptomes for single-cell deciphering},
  author={Shen, Hongru and Liu, Jilei and Hu, Jiani and Shen, Xilin and Zhang, Chao and Wu, Dan and Feng, Mengyao and Yang, Meng and Li, Yang and Yang, Yichen and others},
  journal={Iscience},
  volume={26},
  number={5},
  year={2023},
  publisher={Elsevier}
}

@article{gong2024xtrimogene,
  title={xTrimoGene: an efficient and scalable representation learner for single-cell RNA-seq data},
  author={Gong, Jing and Hao, Minsheng and Cheng, Xingyi and Zeng, Xin and Liu, Chiming and Ma, Jianzhu and Zhang, Xuegong and Wang, Taifeng and Song, Le},
  journal={Advances in Neural Information Processing Systems},
  volume={36},
  year={2024}
}

@article{elnaggar2023ankh,
  title={Ankh: Optimized protein language model unlocks general-purpose modelling},
  author={Elnaggar, Ahmed and Essam, Hazem and Salah-Eldin, Wafaa and Moustafa, Walid and Elkerdawy, Mohamed and Rochereau, Charlotte and Rost, Burkhard},
  journal={arXiv preprint arXiv:2301.06568},
  year={2023}
}

@article{geffen2022distilprotbert,
  title={DistilProtBert: a distilled protein language model used to distinguish between real proteins and their randomly shuffled counterparts},
  author={Geffen, Yaron and Ofran, Yanay and Unger, Ron},
  journal={Bioinformatics},
  volume={38},
  number={Supplement\_2},
  pages={ii95--ii98},
  year={2022},
  publisher={Oxford University Press}
}

@article{rives2021biological,
  title={Biological structure and function emerge from scaling unsupervised learning to 250 million protein sequences},
  author={Rives, Alexander and Meier, Joshua and Sercu, Tom and Goyal, Siddharth and Lin, Zeming and Liu, Jason and Guo, Demi and Ott, Myle and Zitnick, C Lawrence and Ma, Jerry and others},
  journal={Proceedings of the National Academy of Sciences},
  volume={118},
  number={15},
  pages={e2016239118},
  year={2021},
  publisher={National Acad Sciences}
}

@article{meier2021language,
  title={Language models enable zero-shot prediction of the effects of mutations on protein function},
  author={Meier, Joshua and Rao, Roshan and Verkuil, Robert and Liu, Jason and Sercu, Tom and Rives, Alex},
  journal={Advances in neural information processing systems},
  volume={34},
  pages={29287--29303},
  year={2021}
}

@inproceedings{rao2021msa,
  title={MSA transformer},
  author={Rao, Roshan M and Liu, Jason and Verkuil, Robert and Meier, Joshua and Canny, John and Abbeel, Pieter and Sercu, Tom and Rives, Alexander},
  booktitle={International Conference on Machine Learning},
  pages={8844--8856},
  year={2021},
  organization={PMLR}
}

@article{alamdari2023protein,
  title={Protein generation with evolutionary diffusion: sequence is all you need},
  author={Alamdari, Sarah and Thakkar, Nitya and van den Berg, Rianne and Lu, Alex Xijie and Fusi, Nicolo and Amini, Ava Pardis and Yang, Kevin K},
  journal={bioRxiv},
  pages={2023--09},
  year={2023},
  publisher={Cold Spring Harbor Laboratory}
}

@inproceedings{proberta,
  title={Transforming the language of life: transformer neural networks for protein prediction tasks},
  author={Nambiar, Ananthan and Heflin, Maeve and Liu, Simon and Maslov, Sergei and Hopkins, Mark and Ritz, Anna},
  booktitle={Proceedings of the 11th ACM international conference on bioinformatics, computational biology and health informatics},
  pages={1--8},
  year={2020}
}

@article{madani2023large,
  title={Large language models generate functional protein sequences across diverse families},
  author={Madani, Ali and Krause, Ben and Greene, Eric R and Subramanian, Subu and Mohr, Benjamin P and Holton, James M and Olmos, Jose Luis and Xiong, Caiming and Sun, Zachary Z and Socher, Richard and others},
  journal={Nature Biotechnology},
  volume={41},
  number={8},
  pages={1099--1106},
  year={2023},
  publisher={Nature Publishing Group}
}

@article {prostt5,
	author = {Heinzinger, Michael and Weissenow, Konstantin and Sanchez, Joaquin Gomez and Henkel, Adrian and Steinegger, Martin and Rost, Burkhard},
	title = {ProstT5: Bilingual Language Model for Protein Sequence and Structure},
	elocation-id = {2023.07.23.550085},
	year = {2023},
	doi = {10.1101/2023.07.23.550085},
	publisher = {Cold Spring Harbor Laboratory},
	URL = {https://www.biorxiv.org/content/early/2023/07/25/2023.07.23.550085},
	eprint = {https://www.biorxiv.org/content/early/2023/07/25/2023.07.23.550085.full.pdf},
	journal = {bioRxiv}
}

@article{prottrans,
  title={Prottrans: Toward understanding the language of life through self-supervised learning},
  author={Elnaggar, Ahmed and Heinzinger, Michael and Dallago, Christian and Rehawi, Ghalia and Wang, Yu and Jones, Llion and Gibbs, Tom and Feher, Tamas and Angerer, Christoph and Steinegger, Martin and others},
  journal={IEEE transactions on pattern analysis and machine intelligence},
  volume={44},
  number={10},
  pages={7112--7127},
  year={2021},
  publisher={IEEE}
}

@article{ferruz2022protgpt2,
  title={ProtGPT2 is a deep unsupervised language model for protein design},
  author={Ferruz, Noelia and Schmidt, Steffen and H{\"o}cker, Birte},
  journal={Nature communications},
  volume={13},
  number={1},
  pages={4348},
  year={2022},
  publisher={Nature Publishing Group UK London}
}

@article{su2023saprot,
  title={Saprot: Protein language modeling with structure-aware vocabulary},
  author={Su, Jin and Han, Chenchen and Zhou, Yuyang and Shan, Junjie and Zhou, Xibin and Yuan, Fajie},
  journal={bioRxiv},
  pages={2023--10},
  year={2023},
  publisher={Cold Spring Harbor Laboratory}
}

@inproceedings{notin2022tranception,
  title={Tranception: protein fitness prediction with autoregressive transformers and inference-time retrieval},
  author={Notin, Pascal and Dias, Mafalda and Frazer, Jonathan and Marchena-Hurtado, Javier and Gomez, Aidan N and Marks, Debora and Gal, Yarin},
  booktitle={International Conference on Machine Learning},
  pages={16990--17017},
  year={2022},
  organization={PMLR}
}

@article {xtrimo100b,
	author = {Chen, Bo and Cheng, Xingyi and Geng, Yangli-ao and Li, Shen and Zeng, Xin and Wang, Boyan and Gong, Jing and Liu, Chiming and Zeng, Aohan and Dong, Yuxiao and Tang, Jie and Song, Le},
	title = {xTrimoPGLM: Unified 100B-Scale Pre-trained Transformer for Deciphering the Language of Protein},
	elocation-id = {2023.07.05.547496},
	year = {2023},
	doi = {10.1101/2023.07.05.547496},
	publisher = {Cold Spring Harbor Laboratory},
	URL = {https://www.biorxiv.org/content/early/2023/07/14/2023.07.05.547496},
	eprint = {https://www.biorxiv.org/content/early/2023/07/14/2023.07.05.547496.full.pdf},
	journal = {bioRxiv}
}

@article{bert6ma,
    author = {Tsukiyama, Sho and Hasan, Md Mehedi and Deng, Hong-Wen and Kurata,  Hiroyuki},
    title = "{BERT6mA: prediction of DNA N6-methyladenine site using deep learning-based approaches}",
    journal = {Briefings in Bioinformatics},
    volume = {23},
    number = {2},
    pages = {bbac053},
    year = {2022},
    month = {02},
    issn = {1477-4054},
    doi = {10.1093/bib/bbac053},
    url = {https://doi.org/10.1093/bib/bbac053},
    eprint = {https://academic.oup.com/bib/article-pdf/23/2/bbac053/42806331/bbac053.pdf},
}

@article{chromoformer,
  title={Learning the histone codes with large genomic windows and three-dimensional chromatin interactions using transformer},
  author={Lee, Dohoon and Yang, Jeewon and Kim, Sun},
  journal={Nature Communications},
  volume={13},
  number={1},
  pages={6678},
  year={2022},
  publisher={Nature Publishing Group UK London}
}

@article{cpgtransformer,
    author = {De Waele, Gaetan and Clauwaert, Jim and Menschaert, Gerben and Waegeman, Willem},
    title = "{CpG Transformer for imputation of single-cell methylomes}",
    journal = {Bioinformatics},
    volume = {38},
    number = {3},
    pages = {597-603},
    year = {2021},
    month = {10},
    issn = {1367-4803},
    doi = {10.1093/bioinformatics/btab746},
    url = {https://doi.org/10.1093/bioinformatics/btab746},
    eprint = {https://academic.oup.com/bioinformatics/article-pdf/38/3/597/49008701/btab746.pdf},
}

@article{idnaabf,
  title={iDNA-ABF: multi-scale deep biological language learning model for the interpretable prediction of DNA methylations},
  author={Jin, Junru and Yu, Yingying and Wang, Ruheng and Zeng, Xin and Pang, Chao and Jiang, Yi and Li, Zhongshen and Dai, Yutong and Su, Ran and Zou, Quan and others},
  journal={Genome biology},
  volume={23},
  number={1},
  pages={219},
  year={2022},
  publisher={Springer}
}

@article{zhou2022deep,
  title={Deep learning predicts DNA methylation regulatory variants in the human brain and elucidates the genetics of psychiatric disorders},
  author={Zhou, Jiyun and Chen, Qiang and Braun, Patricia R and Perzel Mandell, Kira A and Jaffe, Andrew E and Tan, Hao Yang and Hyde, Thomas M and Kleinman, Joel E and Potash, James B and Shinozaki, Gen and others},
  journal={Proceedings of the National Academy of Sciences},
  volume={119},
  number={34},
  pages={e2206069119},
  year={2022},
  publisher={National Acad Sciences}
}

@article{phosphormer,
    author = {Zhou, Zhongliang and Yeung, Wayland and Gravel, Nathan and Salcedo, Mariah and Soleymani, Saber and Li, Sheng and Kannan, Natarajan},
    title = "{Phosformer: an explainable transformer model for protein kinase-specific phosphorylation predictions}",
    journal = {Bioinformatics},
    volume = {39},
    number = {2},
    pages = {btad046},
    year = {2023},
    month = {01},
    issn = {1367-4811},
    doi = {10.1093/bioinformatics/btad046},
    url = {https://doi.org/10.1093/bioinformatics/btad046},
    eprint = {https://academic.oup.com/bioinformatics/article-pdf/39/2/btad046/49096208/btad046.pdf},
}

@article{randommask,
  title={Toward Understanding BERT-Like Pre-Training for DNA Foundation Models},
  author={Liang, Chaoqi and Qiao, Lifeng and Ye, Peng and Dong, Nanqing and Sun, Jianle and Bai, Weiqiang and Ren, Yuchen and Ma, Xinzhu and Yan, Hongliang and Song, Chunfeng and others},
  journal={arXiv preprint arXiv:2310.07644},
  year={2023}
}

@article{bepler2019learning,
  title={Learning protein sequence embeddings using information from structure},
  author={Bepler, Tristan and Berger, Bonnie},
  journal={International Conference on Learning Representations},
  year={2019}
}

@article{alley2019unified,
  title={Unified rational protein engineering with sequence-based deep representation learning},
  author={Alley, Ethan C and Khimulya, Grigory and Biswas, Surojit and AlQuraishi, Mohammed and Church, George M},
  journal={Nature methods},
  volume={16},
  number={12},
  pages={1315--1322},
  year={2019},
  publisher={Nature Publishing Group US New York}
}

@article{pandey2024deepnap,
  title={DeePNAP: A Deep Learning Method to Predict Protein--Nucleic Acid Binding Affinity from Their Sequences},
  author={Pandey, Uddeshya and Behara, Sasi M and Sharma, Siddhant and Patil, Rachit S and Nambiar, Souparnika and Koner, Debasish and Bhukya, Hussain},
  journal={Journal of Chemical Information and Modeling},
  volume={64},
  number={6},
  pages={1806--1815},
  year={2024},
  publisher={ACS Publications}
}

@inproceedings{jaegle2021perceiver,
  title={Perceiver: General perception with iterative attention},
  author={Jaegle, Andrew and Gimeno, Felix and Brock, Andy and Vinyals, Oriol and Zisserman, Andrew and Carreira, Joao},
  booktitle={International conference on machine learning},
  pages={4651--4664},
  year={2021},
  organization={PMLR}
}

@article{alayrac2022flamingo,
  title={Flamingo: a visual language model for few-shot learning},
  author={Alayrac, Jean-Baptiste and Donahue, Jeff and Luc, Pauline and Miech, Antoine and Barr, Iain and Hasson, Yana and Lenc, Karel and Mensch, Arthur and Millican, Katherine and Reynolds, Malcolm and others},
  journal={Advances in neural information processing systems},
  volume={35},
  pages={23716--23736},
  year={2022}
}

@inproceedings{radford2021learning,
  title={Learning transferable visual models from natural language supervision},
  author={Radford, Alec and Kim, Jong Wook and Hallacy, Chris and Ramesh, Aditya and Goh, Gabriel and Agarwal, Sandhini and Sastry, Girish and Askell, Amanda and Mishkin, Pamela and Clark, Jack and others},
  booktitle={International conference on machine learning},
  pages={8748--8763},
  year={2021},
  organization={PMLR}
}

@article{liu2024visual,
  title={Visual instruction tuning},
  author={Liu, Haotian and Li, Chunyuan and Wu, Qingyang and Lee, Yong Jae},
  journal={Advances in neural information processing systems},
  volume={36},
  year={2024}
}

@article{JASPAR,
  title={JASPAR 2024: 20th anniversary of the open-access database of transcription factor binding profiles},
  author={Rauluseviciute, Ieva and Riudavets-Puig, Rafael and Blanc-Mathieu, Romain and Castro-Mondragon, Jaime A and Ferenc, Katalin and Kumar, Vipin and Lemma, Roza Berhanu and Lucas, J{\'e}r{\'e}my and Ch{\`e}neby, Jeanne and Baranasic, Damir and others},
  journal={Nucleic acids research},
  volume={52},
  number={D1},
  pages={D174--D182},
  year={2024},
  publisher={Oxford University Press}
}

@article{lucaone,
  title={LucaOne: generalized biological foundation model with unified nucleic acid and protein language},
  author={He, Yong and Fang, Pan and Shan, Yongtao and Pan, Yuanfei and Wei, Yanhong and Chen, Yichang and Chen, Yihao and Liu, Yi and Zeng, Zhenyu and Zhou, Zhan and others},
  journal={bioRxiv},
  pages={2024--05},
  year={2024},
  publisher={Cold Spring Harbor Laboratory}
}

@article{blosum62,
  title={Amino acid substitution matrices from protein blocks.},
  author={Henikoff, Steven and Henikoff, Jorja G},
  journal={Proceedings of the National Academy of Sciences},
  volume={89},
  number={22},
  pages={10915--10919},
  year={1992}
}

@inproceedings{clip,
  title={Learning transferable visual models from natural language supervision},
  author={Radford, Alec and Kim, Jong Wook and Hallacy, Chris and Ramesh, Aditya and Goh, Gabriel and Agarwal, Sandhini and Sastry, Girish and Askell, Amanda and Mishkin, Pamela and Clark, Jack and others},
  booktitle={International conference on machine learning},
  pages={8748--8763},
  year={2021},
  organization={PmLR}
}

@article{wetlabdeltaG,
  title={The experimental uncertainty of heterogeneous public K i data},
  author={Kramer, Christian and Kalliokoski, Tuomo and Gedeck, Peter and Vulpetti, Anna},
  journal={Journal of medicinal chemistry},
  volume={55},
  number={11},
  pages={5165--5173},
  year={2012},
  publisher={ACS Publications}
}

@article{drori2019accurate,
  title={Accurate protein structure prediction by embeddings and deep learning representations},
  author={Drori, Iddo and Thaker, Darshan and Srivatsa, Arjun and Jeong, Daniel and Wang, Yueqi and Nan, Linyong and Wu, Fan and Leggas, Dimitri and Lei, Jinhao and Lu, Weiyi and Fu, Weilong and others},
  journal={arXiv preprint arXiv:1911.05531},
  year={2019}
}

@article{pruitt2020refseq,
  title={RefSeq frequently asked questions (FAQ)},
  author={Pruitt, Kim and Murphy, Terence and Brown, Garth and Murphy, Mike},
  journal={RefSeq Help. National Center for Biotechnology Information (US)},
  year={2020}
}

@article{refseq,
  title={Reference sequence (RefSeq) database at NCBI: current status, taxonomic expansion, and functional annotation},
  author={O'Leary, Nuala A and Wright, Mathew W and Brister, J Rodney and Ciufo, Stacy and Haddad, Diana and McVeigh, Rich and Rajput, Bhanu and Robbertse, Barbara and Smith-White, Brian and Ako-Adjei, Danso and others},
  journal={Nucleic acids research},
  volume={44},
  number={D1},
  pages={D733--D745},
  year={2016},
  publisher={Oxford University Press}
}

@article{mbert,
  title={Are all languages created equal in multilingual BERT?},
  author={Wu, Shijie and Dredze, Mark},
  journal={arXiv preprint arXiv:2005.09093},
  year={2020}
}

@misc{su2023roformerenhancedtransformerrotary,
  title={Roformer: Enhanced transformer with rotary position embedding},
  author={Su, Jianlin and Ahmed, Murtadha and Lu, Yu and Pan, Shengfeng and Bo, Wen and Liu, Yunfeng},
  journal={Neurocomputing},
  volume={568},
  pages={127063},
  year={2024},
  publisher={Elsevier}
}

\end{document}